\documentclass[twocolumn]{article}
\usepackage[scale=0.8]{geometry}

\usepackage{authblk}  
\usepackage{natbib}

\usepackage{times}  % DO NOT CHANGE THIS
\usepackage{helvet}  % DO NOT CHANGE THIS
\usepackage{courier}  % DO NOT CHANGE THIS
\usepackage[hyphens]{url}  % DO NOT CHANGE THIS
\usepackage{graphicx} % DO NOT CHANGE THIS
\usepackage{natbib}  % DO NOT CHANGE THIS AND DO NOT ADD ANY OPTIONS TO IT
\usepackage{caption} % DO NOT CHANGE THIS AND DO NOT ADD ANY OPTIONS TO IT
\usepackage{algorithm}
\usepackage{algorithmic}

\usepackage{newfloat}
\usepackage{listings}

\usepackage[hidelinks]{hyperref}

\usepackage{comment}
\usepackage{xcolor}
\usepackage{tikz}
\usepackage{multirow}
\usepackage{subfig}
\usepackage{graphicx}
\usepackage{amsmath}
\usepackage{amssymb}
\usepackage{mathtools}
\usepackage{amsthm}
\usepackage{algorithm}
\usepackage{algorithmic}

\newcommand{\revision}[1]{\textcolor{black}{#1}}

\DeclareMathOperator*{\argmax}{argmax}

\usepackage{todonotes}

\DeclareCaptionStyle{ruled}{labelfont=normalfont,labelsep=colon,strut=off} % DO NOT CHANGE THIS
\floatstyle{ruled}
\newfloat{listing}{tb}{lst}{}
\floatname{listing}{Listing}
\title{Data Heterogeneity and Forgotten Labels in Split Federated Learning\thanks{This manuscript has been accepted for publication in  \textbf{AAAI 2026.}}}

\author[1]{Joana Tirana\footnote{Work completed while at Telef\'onica Scientific Research.}}
\author[2]{Dimitra Tsigkari}
\author[2]{David Solans Noguero}
\author[3]{Nicolas Kourtellis}
\affil[1]{University College Dublin, Ireland}
\affil[2]{Telef\'{o}nica Scientific Research, Barcelona, Spain}
\affil[3]{Keysight AI Labs, Barcelona, Spain}

\date{}

\begin{document}

\maketitle

\begin{abstract}
  In Split Federated Learning (SFL), the clients collaboratively train a model with the help of a server by splitting the model into two parts. Part-1 is trained locally at each client and aggregated by the aggregator at the end of each round. Part-2 is trained at a server that sequentially processes the intermediate activations received from each client. We study the phenomenon of catastrophic forgetting~(CF) in SFL in the presence of data heterogeneity. In detail, due to the nature of SFL, local updates of part-1 may drift away from global optima, while part-2 is sensitive to the processing sequence, similar to forgetting in continual learning~(CL). Specifically, we observe that the trained model performs better in classes~(labels) seen at the end of the sequence. We investigate this phenomenon with emphasis on key aspects of SFL, such as the processing order at the server and the cut layer. Based on our findings, we propose Hydra, a novel mitigation method inspired by multi-head neural networks and adapted for the SFL setting. Extensive numerical evaluations show that Hydra outperforms baselines and methods from the literature.
\end{abstract}

%
\begin{comment}
 \begin{links}
     \link{Code}{https://github.com/jtirana98/Hydra-CF-in-SFL}
    \link{Extended version}{https://aaai.org/example/extended-version}
 \end{links}
\end{comment}

\section{Introduction}
\label{sec:intro}

Split Learning~(SL) is a distributed learning method where part of the clients' training is offloaded to a server~\citep{vepakomma2018split}. 
This is particularly useful in cases where clients' resources are insufficient to perform on-device training. 
In a nutshell, in SL, the deep 
neural network~(NN) model is split into two parts of consecutive layers: part-1 and part-2, where the last layer of part-1 is called the cut layer. 
These parts are trained by the client
and the server, respectively.
There is a plethora of proposed variations for SL
that mostly differ in the way the training at the server takes place. 
In this work, we focus on 
Split Federated Learning~(SFL) and, in particular, SplitFedv2~\citep{thapa2022splitfed}, which is one of the most prevailing versions~\citep{hafi2024split}.
In fact, thanks to its workflow, SFL has been shown to deliver a faster convergence than other variants~\citep{thapa2022splitfed}.

\begin{figure}[t!]
    \centering
\includegraphics[clip, width=0.83\columnwidth, trim={0.88cm 13.7cm 2.7cm 1.54cm}]{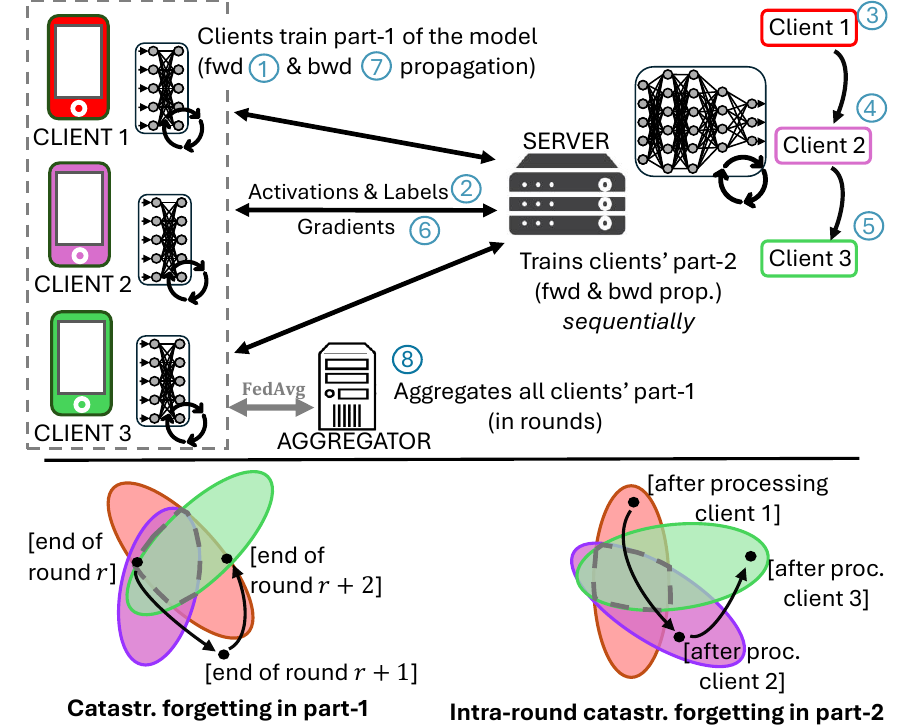}
\caption{(Top) The steps of the processing workflow of SFL.
(Bottom) Parameter spaces for low error for clients’ dominant labels 
(under data heterogeneity),
where a client's color represents its dominant label. Part-1 of the model suffers from catastrophic forgetting from round to round.
Part-2 suffers from intra-round catastrophic forgetting due to the processing order at the server.
}
\label{fig:intro_pic}
\end{figure}

The processing workflow of SFL for 3 clients is depicted in Fig.~\ref{fig:intro_pic} (top). At the beginning of every round, the clients perform forward propagation of part-1 of the model on a batch of their samples (step 1 in Fig.~\ref{fig:intro_pic}) and transmit the activations of the cut layer to the server, along with the labels of the batch~(step 2). Then, the server sequentially trains part-2 for each client, propagating forward and backward~(steps 3-5),  and transmits the gradients to the clients~(step 6). 
Based on the received gradients, the clients perform backward propagation of part-1 (step 7). Steps 1 to 7 are repeated for all the batches, and once all clients have processed all their data, the aggregator aggregates the local version of part-1 and sends a global part-1 to the clients~(step 8), before a new round begins. We consider here the typical case where clients' data remains 
unchanged throughout the training.

\begin{figure}[t!]
    \centering
\includegraphics[clip, width=1
\columnwidth, trim={0cm 0.2cm 0cm 0.1cm}]{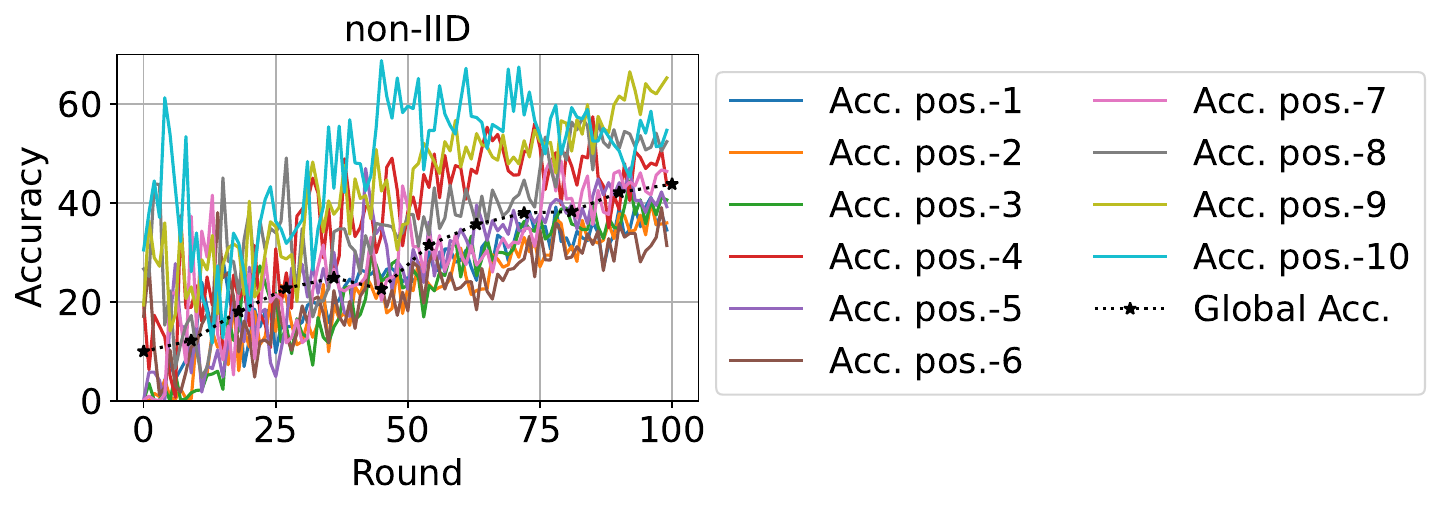}
\caption{Accuracy per processing position (at the server) and global accuracy achieved by MobileNet in CIFAR-10 under non-IID data distributions among $10$ clients. 
}
\label{fig:intro_nonIID}
\end{figure}

\noindent \textbf{Motivation.} Part-1 is trained as in Federated Learning~(FL), while part-2 is trained over the activations that clients send. This \emph{dual aspect} of  SFL makes it particularly 
susceptible to catastrophic forgetting~(CF) in the presence of data heterogeneity, a common phenomenon in training 
NNs~\citep{goodfellow2013empirical}.
In practice, part-1 and part-2 may suffer from CF differently, as depicted in Fig.~\ref{fig:intro_pic}~(bottom). Like in FL, the local updates of part-1 drift away from the global optima as a result of the aggregation~(e.g., \texttt{FedAvg}~\citep{mcmahan2017communication}) 
at the end of every round. On the other hand, our experiments reveal a new finding: the processing order at the server has a significant impact on CF in part-2. 
This CF resembles the one of Continual Learning~(CL)~\citep{wang2024comprehensive}: the model parameters perform well for the dominant label
of the last processed client, but forget the previously seen labels, i.e., in every update step at the server, part-2 tends to forget the knowledge learned during the previous steps, leading to a \emph{performance gap} among labels.

Despite the resemblance 
to forgetting in CL,
there are several differences between SFL and CL. Notably, in SFL, the data distribution is fixed and the data stream is finite, i.e., the data is not temporal. Further, two types of repetitions occur~(not to be confused with knowledge replay): (i) all clients' input (i.e., part-1's activations) is cyclically passed through the server at each round, and (ii) depending on the processing order of the clients' part-2, input concerning a specific label may be repeated within the same round.

Fig.~\ref{fig:intro_nonIID} showcases the CF in SFL as a result of the processing order at the server under non-independently and identically
distributed (non-IID) data.  
In particular, we assume that the data of each client contains a highly represented~(dominant) label. 
Each line depicts the accuracy of the labels that are dominant in clients that are processed at a certain position (1 to 10, in  CIFAR-10 with 10 labels), where details on the experimental setup will be  given in Sec.~\ref{sec:method_exper}. We see that, under non-IID data, the labels that are dominant at the clients who are processed last at the server consistently outperform the other labels. 
The disparity in performance between the labels seen at the end of the sequence and those seen earlier is related to the CF in part-2. 
We identify this as \emph{intra-round catastrophic forgetting}~(intraCF), as it occurs at a different granularity (i.e., within a round) than the model drifting of part-1, which also contributes to the phenomenon. 
Moreover,  the global accuracy is heavily affected, i.e., $43\%$ after 100 rounds, whereas, under IID data, the global accuracy reaches $80\%$ after 100 rounds.

\noindent \textbf{Challenges.} Despite the plethora of works on CF in other settings (CL, FL) and some recent 
findings
in SL and SFL (further discussed in Sec.~\ref{sec:related_work}), we identify several challenges.\\
\noindent \textbf{(i)} \emph{Metrics of forgetting \& Mitigation Methods.} Related work on forgetting in SL, e.g.,  \citep{feng2023iotsl, xia2024korea} focuses solely on  the global accuracy, ignoring the disparity (gap) of performance among labels. 
Further, methods proposed to mitigate forgetting and/or data heterogeneity in CL and FL, such as EWC~\citep{kirkpatrick2017overcoming} and Scaffold~\citep{karimireddy2020scaffold}, cannot be applied in the setting of SFL as the former is based on the notion of temporal data and the latter requires the entire model. \\
\noindent \textbf{(ii)} \emph{The roles of the cut layer and processing order.} The majority of related work assumes that the server processes the clients' data in a random order~\citep{thapa2022splitfed, liao2023mergesfl} and often ignores the role of the cut layer in CF. Moreover, existing theoretical analyses on SFL are limited to convergence analysis under standard assumptions without focusing on these key elements of SFL~\citep{hanconvergence}. \\
\noindent \textbf{(iii)}  \emph{Empirical evaluation.} The existing theoretical findings are often showcased in limited evaluation scenarios~(e.g., partition methods),
 see~\citep{li2023convergence, hanconvergence}. 
Further, it is unclear whether some of the theoretical claims proved in the settings of CL and FL may hold in SFL.

\noindent \textbf{Contributions.} The challenges above further corroborate the need for an empirical evaluation of the phenomenon of catastrophic forgetting in SFL and for an efficient mitigation method. This manuscript addresses these challenges  by:\\
\noindent \textbf{1.} Identifying a new case of CF in SFL 
as a result of the processing workflow of SFL. \\
\noindent \textbf{2.} Providing valuable insights into this phenomenon (e.g., w.r.t. the impact of cut layer and processing order at the server) and comparing them with existing theoretical findings.
To this end, we focus on the accuracy, the performance gap among labels, and the metric of backward transfer. \\
\noindent \textbf{3.} Proposing Hydra, a novel mitigation method that trains multiple versions of the last layers of part-2 after grouping the clients' input based on their data distributions.
Hydra is designed based on the insights from our analysis and induces minimal overhead~(which can be further reduced).
\\
\textbf{4.} Showing how Hydra reduces CF and outperforms other methods through numerical evaluations in a variety of NN models, datasets, and data partitions.

\emph{Our implementation code is publicly available at:
https://github.com/jtirana98/Hydra-CF-in-SFL.}

\section{Related Work}
\label{sec:related_work}

In this section, we discuss related work on data heterogeneity and CF in FL and SL. In FL under non-IID data, several works study CF as a result of model drifting and client selection policies. Mitigation methods utilize knowledge distillation~\citep{lee2022preservation, ma2022continual} and regularization~\citep{xu2022acceleration}, among others. We notice, however, that most of the methods for FL require control of the entire model, e.g., \citep{zhao2018federated, karimireddy2020scaffold,hu2024fedmut}, which is incompatible with the SFL workflow. Moreover, optimization techniques in FL~\citep{li2020federated, wang2020tackling} assume a symmetric training workflow with full local steps or per-client regularization terms, which  is also not applicable for SFL. On the other hand, replay-based methods for FL, e.g., \citep{qibetter,li2024towards} induce computational and memory overhead.
Concerning SL (in its various definitions), CF has been mostly studied in terms of global accuracy under non-IID data. Proposed solutions are based on generated data and knowledge transfer for SplitNN~\citep{feng2023iotsl, feng2024slwf}, as well as on
knowledge replay for SplitFedv1~\citep{xia2024korea}. Further,~\citep{madaan2022vulnerability} provides a preliminary study of the impact of the processing order in different variants of SL. Finally, the only work that addresses data heterogeneity but not CF in SFL~(SplitFedv2) is~\citep{liao2023mergesfl}, which merges features and adjusts the batch size.

\section{Exploring CF in SFL}
\label{sec:intraCF}

The goal is to train a (deep) classification model whose set of labels~(classes) is $\mathcal{L}$, where $|\mathcal{L}|=L$. The training with a set $\mathcal{C}$ of clients follows the SFL workflow described in Sec.~\ref{sec:intro}. Further, clients' data can be seen as ``tasks" (in the parlance of CL) with possibly (totally) overlapping labels.

\subsection{Methodology of Experiments}
\label{sec:method_exper}

\textbf{Models, Datasets, and Cut Layer in SFL.} Two ML models are employed: MobileNetV1~\citep{howard2017mobilenets}, known for its efficiency on mobile and resource-constrained devices,   and ResNet101~\citep{he2016deep}, which is renowned for its depth and robust feature extraction capabilities. 
These models are 
trained on SVHN~\citep{netzer2011reading}, CIFAR-10~($L=10$), CIFAR-100~\citep{krizhevsky2009learning}~($L=20$ or $100$ -- unless otherwise specified, we use $20$ superclasses as in~\citep{ramasesh2020anatomy}), and TinyImageNet~\citep{le2015tiny}~($L=200$).
The cut layer
determines the sizes of part-1 and part-2. A shallow cut layer implies a small part-1, i.e., the server trains the model's largest portion, while a deep one implies a small part-2. 

Based on the discussion in Sec.~\ref{sec:intro} about the intraCF,  unless otherwise specified, we choose a shallow cut layer, e.g., at layer 4 (out of 26 in  total) in MobileNetV1, and layer 2 (out of 35 in total) in ResNet101, as is common in related work~\citep{thapa2022splitfed}. In fact, in Sec.~\ref{sec:intraCF}, we show that a shallow cut better captures the impact of intraCF on the forgetting observed in SFL.
We stress, however, that we also perform a sensitivity analysis with respect to the choice of the cut layer, while following the bottleneck approach~\citep{kang2017neurosurgeon}. Finally, all the experiments are repeated at least $10$ times for $100$ training rounds~(with varying random seeds) and, then, for each metric (e.g., global accuracy), the median 
of the observed values from all the repeated experiments is computed. This is to ensure that our results are robust against outliers~\citep{leys2013detecting}.

\noindent \textbf{Data Partitioning and Number of Clients.} 
We use three common data partitioning techniques for data heterogeneity. 
First, we employ the \textit{Dominant Label~(DL) ratio} method, an instance of overlapping label distribution techniques~\citep{solans2024non}. 
This uses the parameter~$p \in [0,100]$ to control the percentage~(i.e., $p\%$) of samples with a dominant label at each client, while the remaining samples~(i.e., $(100-p)\%$) are distributed evenly among the other clients. Clearly, as $p$ increases, greater heterogeneity is produced.
Throughout the paper, the notation $p\%$-DL is employed, and we use the parameter~$\phi\in \mathbb{N}$ to describe the number of clients that, for each label, have this label as dominant. Hence, the total number of clients participating in the training is $C=|\mathcal{C}| = \phi \cdot L$. We also employ the \textit{Sharding} and \textit{Dirichlet} data partitioning methods.

\noindent \textbf{Metrics for Catastrophic Forgetting in SFL.} A common metric of forgetting in CL is the \emph{backward transfer}~(BW), which has been adapted for FL as follows~\citep{lee2022preservation}:
\begin{equation}
\mathcal{BW} = \frac{1}{L} \sum_{l=1}^{L} \max_{r\in \{1, \ldots, R-1\}} (A_l^{r}-A_l^{R}),
\label{eq:forgetting_score}
\end{equation}  
where $A_l^{r}$ is the accuracy of label $l$ at round $r < R$. It captures the difference between the maximum (``peak") accuracy 
and the final accuracy of each label at the end of training (i.e., round $R$), averaged over all labels. The BW metric is better suited for \textit{dynamic input data}, where the performance of a label may decrease with time as other labels appear more often (e.g., in temporal data). For this reason, and motivated by what we observe in Fig.~\ref{fig:intro_nonIID}, we also measure the (average) \emph{performance gap~(PG)} between labels, for each round $r$, as follows:

\begin{equation}
\mathcal{PG}(r) = \frac{1}{L} \sum_{l=1}^{L} \max_{k \in \mathcal{L}} \{ |\min(0, (A_l^r - A_{k}^r))|\}.
\label{eq:forgetting_score_flashback}
\end{equation}
The quantity above measures the maximum difference in performance between each label and the best-performing one, averaged over all labels. Essentially, $\mathcal{PG}(r)$ captures the average ``gap" between the per-label accuracies at round~$r$.

We evaluate CF with the two metrics above, while, in Sec.~\ref{sec:exp}, we employ an additional metric in the case of cyclic order and DL partitioning, the \emph{per-position accuracy}. In practice, position-$k$ accuracy, for $k\in \{1 \ldotp \ldotp L\}$, is the accuracy of the labels that are dominant in clients that are processed at position $k$~(in the cycle of length $L$) at the server.

\subsection{Insights into Catastrophic Forgetting in SFL}
\label{sec:order}

\begin{table*}[!t]  
\centering 
\begin{tabular}{p{1.35cm}p{0.5cm}|p{0.8cm}p{0.8cm}p{1cm}p{0.9cm}|p{0.7cm}p{0.8cm}p{1.cm}p{0.9cm}|p{0.8cm}p{0.7cm}p{0.7cm}p{0.8cm}}

& \small{\textbf{DL}} & \multicolumn{4}{c|}{\small{\textbf{Global Accuracy} ($\uparrow$)}}  & \multicolumn{4}{c|}{\small{\textbf{Performance Gap (PG)} ($\downarrow$)}}  & \multicolumn{4}{c}{\small{\textbf{Backward Transfer (BW)} ($\downarrow$)}} \\
\small{\textbf{Model}} & \small{\textbf{ratio}} & \small{$\phi_{=1}$} & \small{$\phi_{=5}$}& \small{$\phi_{=10}$} & \small{\textbf{random}} & \small{$\phi_{=1}$} & \small{$\phi_{=5}$} & \small{$\phi_{=10}$} & \small{\textbf{random}} & \small{$\phi_{=1}$} & \small{$\phi_{=5}$} & \small{$\phi_{=10}$} & \small{\textbf{random}}\\ \hline %\hline 
 & & \multicolumn{12}{c}{\small{\textit{\textbf{CIFAR-10}}}} \\ 
\multirow[c]{1}{*}{MobileNet}  & $80$  & $\mathbf{45}$\scriptsize{$\mathbf{\pm 0.8}$} & $43.7$\scriptsize{$\pm 1$}  & $41.1$\scriptsize{$\pm 1$} & $35$\scriptsize{$\pm 4$} 
& $\mathbf{29}$\scriptsize{$\mathbf{\pm 3}$} & $38.7$\scriptsize{$\pm 3$} & $43.4$\scriptsize{$\pm 3.6$} & $46.4$\scriptsize{$\pm 3$} &
$33.5$\scriptsize{${\pm 6}$} & $39$\scriptsize{${\pm 6}$} & $\mathbf{28}$\scriptsize{$\mathbf{\pm 5}$} & $56$\scriptsize{$\pm 7$} \\ \cline{2-14} 
\scriptsize{IID:(80, 10)}
& $60$  &  $\mathbf{70}$\scriptsize{$\mathbf{\pm 0.1}$} & $67$\scriptsize{$\pm 0.5$} & $67.5$\scriptsize{$\pm 0.4$}  & $66$\scriptsize{$\pm 2$} &$\mathbf{15}$\scriptsize{$\mathbf{\pm 0.7}$} & $17$\scriptsize{$\pm 0.7$} & $21.6$\scriptsize{$\pm 3.6$}  & $23.4$\scriptsize{$\pm 2$} & 

$10.7$\scriptsize{${\pm 1}$} & $11$\scriptsize{$\pm 2$} & $\mathbf{8}$\scriptsize{$\mathbf{\pm 2}$}  & $27$\scriptsize{${\pm 7}$}  
\\ \hline 
\multirow[c]{1}{*}{ResNet101}  & $80$ & $36.2$\scriptsize{$\pm 1$} &  $\mathbf{38}$\scriptsize{$\mathbf{\pm 0.7}$}  & $21.3$\scriptsize{$\pm 3$} &  $33$\scriptsize{$\pm 3.5$} & $\mathbf{24}$\scriptsize{$\mathbf{\pm 3}$}   & $35.2$\scriptsize{$\pm 3$}  &  $69.7$\scriptsize{$\pm 5$} & $57$\scriptsize{$\pm 5$} &   
$40$\scriptsize{${\pm 10}$} & $29$\scriptsize{${\pm 4}$} & $\mathbf{21}$\scriptsize{$\mathbf{\pm 9}$} & $68$\scriptsize{${\pm 9}$} \\ \cline{2-14}
\scriptsize{IID:(68,15)}
 & $60$   &  $\mathbf{52.4}$\scriptsize{$\mathbf{\pm 1}$} & $51$\scriptsize{$\pm 0.8$} & $45.7$\scriptsize{$\pm 2$} & $48$\scriptsize{$\pm 3$} 
 &  $\mathbf{19}$\scriptsize{$\mathbf{\pm 1}$} & $ 20.6$\scriptsize{$\pm 3$} & $38.2$\scriptsize{$\pm 4.5$}  & $32$\scriptsize{$\pm 6$}  &
$23$\scriptsize{${\pm 4}$} & $22$\scriptsize{$\pm 11$} & $\mathbf{14.5}$\scriptsize{$\mathbf{\pm 1}$}  & $ 47$\scriptsize{$\pm 7$}  
\\ \hline %\hline
& & \multicolumn{12}{c}{\small{\textbf{\textit{CIFAR-100 with $\mathbf{20}$ superclasses ($\mathbf{L=20}$)}}}} \\ 

MobileNet \scriptsize{IID:(44,28)} & $80$  & $36$\scriptsize{$\pm 0.7$} & $36$\scriptsize{$\pm 0.4$} & $27.5$\scriptsize{$\pm 0.3$} & $\mathbf{38}$\scriptsize{$\mathbf{\pm 0.4}$} 
& $31$\scriptsize{$\pm 1.7$} & $\mathbf{29}$\scriptsize{$\mathbf{\pm 1.6}$} & $32$\scriptsize{$\pm 3$} & $31.1$\scriptsize{$\pm 2$} & ${17}$\scriptsize{${\pm 2}$} & $16$\scriptsize{${\pm 1}$} & $\mathbf{17}$\scriptsize{$\mathbf{\pm 3}$} & $38$\scriptsize{${\pm 2}$}
\\ \hline
ResNet101 \scriptsize{IID:(38,29)} & $80$  & $ 23.6$\scriptsize{$\pm 1$} & $20$\scriptsize{$\pm 1.7$} & $\mathbf{26}$\scriptsize{$\mathbf{\pm 2.4}$} & $23$\scriptsize{$\pm 1.6$} & $42$\scriptsize{$\pm 4$} & $\mathbf{40.4}$\scriptsize{$\mathbf{\pm 6}$} & $54.1$\scriptsize{$\pm 2.2$} & $45.6$\scriptsize{$\pm 5$}  & $39$\scriptsize{${\pm 6}$} & $42$\scriptsize{${\pm 7}$} & $\mathbf{25}$\scriptsize{$\mathbf{\pm 5}$} & $60$\scriptsize{${\pm 8}$} \\ \hline %\hline
& & \multicolumn{12}{c}{\small{\textbf{\textit{CIFAR-100 with all classes ($\mathbf{L=100}$)}}}}\\ 
MobileNet \scriptsize{IID:(33, 42)} & $80$
& $27$\scriptsize{$\pm 0.1$} &  $27$\scriptsize{$\pm 0.1$}  &  $28$\scriptsize{$\pm 0.2$}  &  $\mathbf{28}$\scriptsize{$\mathbf{\pm 0.1}$} & $48.6$\scriptsize{$\pm 1$} &  $49$\scriptsize{$\pm 0.8$} &  $\mathbf{46}$\scriptsize{$\mathbf{\pm 1}$} &  $46.7$\scriptsize{$\pm 2$} &  $13$\scriptsize{${\pm 2}$} &  $13$\scriptsize{${\pm 2}$} &  $\mathbf{11}$\scriptsize{$\mathbf{\pm 1}$} &  $22$\scriptsize{${\pm 0.5}$} \\ \hline

& & \multicolumn{12}{c}{\small{\textit{\textbf{SVHN}}}} \\ 
MobileNet \scriptsize{IID:(89,5)} & $80$   & $77.4$\scriptsize{$\pm 0.4$} & $77.7$\scriptsize{$\pm 0.4$} & $\mathbf{78}$\scriptsize{$\mathbf{\pm 0.5}$} & $77$\scriptsize{$\pm 0.6$} & $24$\scriptsize{$\pm 0.3$}  & $24$\scriptsize{$\pm 0.4$} & $\mathbf{22}$\scriptsize{$\mathbf{\pm 0.2}$} & $24$\scriptsize{$\pm 0.7$} &  $\mathbf{6}$\scriptsize{$\mathbf{\pm 1}$} &  $7.7$\scriptsize{${\pm 1}$} &  $7$\scriptsize{${\pm 0.5}$} &  $9$\scriptsize{${\pm 1}$}\\ \hline

ResNet101 \scriptsize{IID:(88,6.2)} & $80$ &  
$63$\scriptsize{$\pm 2$}  & $\mathbf{67}$\scriptsize{$\mathbf{\pm 3}$} & $66.2$\scriptsize{$\pm 1.6$} & $59.7$\scriptsize{$\pm 2$}  & $\mathbf{40}$\scriptsize{$\mathbf{\pm 2}$} & $46$\scriptsize{$\pm 3$} & $42$\scriptsize{$\pm 1$} & $54.9$\scriptsize{$\pm 1$} & $7.7$\scriptsize{${\pm 4}$} &  $\mathbf{6}$\scriptsize{$\mathbf{\pm 3}$} &  $8$\scriptsize{${\pm 5}$} &  $11$\scriptsize{${\pm 6}$}\\
\hline
\end{tabular}
\caption{Global accuracy, Performance Gap, and Backward Transfer (reported median of the last five rounds across all runs) for cyclic \& random order~($\phi$ is the scale of the clients in the cyclic order). The parenthesis under the model type shows the global accuracy and the PG for IID data. The upward/downward arrows indicate that large and small values are desirable, respectively.}
\label{tab:unfairscore_cyclic}
\end{table*}

In what follows, we study CF in SFL, 
while occasionally focusing on intraCF.
We first introduce two different types of processing orders at the server.

\noindent $\bullet$  \textbf{No Specific Order (random).} 
It is often implied in the literature that the server processes the data received by each client in a random or a first-come-first-served~(FCFS) order~\citep{thapa2022splitfed}, %where 
as the arrival time may depend on network
conditions~\citep{tirana2024workflow}. To this end,  we simulate the case of FCFS with a \textit{random} function at each round. \\
$\bullet$ \textbf{Cyclic Order.} 
This order guarantees \emph{structure} in the way the server processes the input of different clients (at part-2). Similar to~\citep{swartworth2023nearly}, at each round, the server processes the clients' input in a cycle of length $L$ 
based on the label that is highly represented in their data, i.e.,  first all $\phi$ clients whose highly represented label is $X\in \mathcal{L}$,
then label $Y\in \mathcal{L}$, and so on. This sequence remains the same across all rounds. 
In our experiments, to ensure that the results are independent of the type of label, the order~(sequence) of the labels is randomly selected in each experiment.

\noindent \textbf{Scale of clients ($\phi$)}. In the cyclic order, larger values of $\phi$ imply that the data of more clients of the same type (e.g., with the same dominant label in DL partition) are processed one after the other.
%, potentially improving global accuracy. 
In other words, $\phi$  gives the scale of repetition of clients of the same type. In the random order, while such repetitions may occur, 
%they are not consistent and 
they may vary from round to round, and thus, $\phi$ is meaningful only in the cyclic order.

Table~\ref{tab:unfairscore_cyclic} shows the global accuracy, PG, and BW for the random and cyclic order with different values of $\phi$ in a variety of models and datasets.
We observe that, in most of the lines, \emph{the best accuracy, PG, and BW are achieved by the cyclic order (for any $\phi$)}, with a few corner cases where random has slightly better accuracy than cyclic, but higher PG.  Also, we observe that among the different values of $\phi$, lower values yield the best accuracy, indicating that diversity in the training sequence is crucial. This finding aligns with other studies in the literature on CL, e.g.,~\citep{lin2023theory}, but has not been showcased in SFL, until now. When $\phi$ is large, the server consecutively processes $\phi$ clients' data that have the same dominant label. This leads to more frequent updates related to a single label, allowing the model to learn this label better, but at the cost of forgetting previous ones.\footnote{A small exception is observed in SVHN, which appears to be less sensitive to $\phi$ because its label distribution is uneven; thus, labels with more samples dominate model updates regardless of $\phi$.} Even though training seems to be more stable~(i.e., lower BW), this leads to an increased PG, which implies higher disparity among labels' performance~(leading to intraCF).  

Both the PG and the BW metrics show that structured processing order policies (at the server), such as the cyclic order, lead to lower forgetting and better training stability when compared to unstructured ones, e.g., random order. This aligns with existing theoretical results on CL and linear regression~\citep{evron2022catastrophic}. Moreover, as we can already  observe in Table~\ref{tab:unfairscore_cyclic} (in terms of accuracy and PG),  many characteristics of forgetting in CL also occur in SFL~(e.g., that larger and more complex models are more sensitive to CF~\citep{ramasesh2021effect}). 
Hence, despite the differences in the two settings~(CL and SFL), forgetting in these settings  carries similarities, which can help to better understand   CF and intraCF in SFL.

\begin{figure}[tb]
    \centering
\includegraphics[width=1\columnwidth]{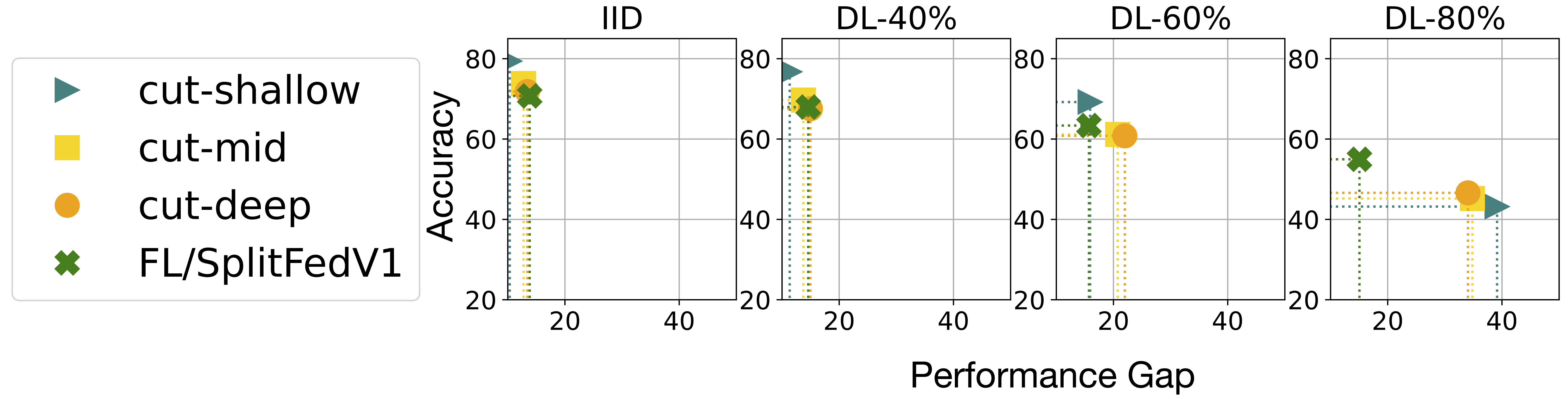}
\caption{Accuracy~(y-axis) and PG~(x-axis) achieved by MobileNet/CIFAR-10 for SFL with different cuts ($4$, $15$, and $23$, resp.)  and cyclic order ($\phi=10$), and FL/SplitFedV1.}
\label{fig:per_position_main}
\end{figure}

\noindent \textbf{Cut Layer and IntraCF.} 
Fig.~\ref{fig:per_position_main} depicts the global accuracy and the PG for different cut layers (in SFL), and for SplitFedv1~\citep{thapa2022splitfed}, a variant of SL. In SplitFedv1,
part-1 is trained as in SFL, and part-2 is trained at the server using a copy for each client, while all copies are aggregated at the end of each round. It has been shown that its performance is the same as that of FL~\citep{gao2021evaluation, thapa2022splitfed}, and hence, we also denote it by FL.
For high data heterogeneity, i.e., $80\%$-DL, as we go from shallow to deep cut layer,  the accuracy increases and the performance gap drops. We recall that the deeper the cut layer is, the larger part-1 becomes, i.e., more layers are trained at the clients (in parallel) and aggregated at the end of the round. 
Hence, \textit{a deep cut layer highlights the effect of CF in part-1, while a shallow cut layer highlights the effect of intraCF on CF.} The former is related to the aggregation under data heterogeneity, while the latter to the processing order at the server. These results imply that the impact of CF in part-1~(model drifting) is smaller than the effect of CF in part-2~(intraCF), i.e., \textit{the cut layer balances the impact of model drifting of part-1 and intraCF of part-2}.

Due to the previous observation, we expect that the deeper the cut layer is in SFL, the more similar the model performs to FL. 
We see that, in terms of accuracy, SFL performs better than FL under IID data. As data heterogeneity increases, however, FL performs better ($54\%$ vs $43\%$, under $80\%$-DL), as also observed by~\citep{gao2020end}, as a result of aggregating the entire model. In terms of CF, under $80\%$-DL, there is a significant performance gap between FL and SFL, for any cut layer. However, we see something unexpected:  the PG of SFL with shallow cut layer~(which showcases intraCF) is very close to that of deep cut layer~(which showcases the CF of part-1). This means that although FL performs better under high heterogeneity, implying that aggregation has a positive impact on CF, the fact that \emph{the last layers of the model are processed at the server (sequentially) leads to high CF in SFL}. 
The importance of the last layers of the model to the training and CF has
also been shown in  CL
in~\citep{ramasesh2020anatomy, zilly2021plasticity}.

\section{Hydra: A Novel Method to Mitigate CF}
\label{sec:medusa}

We observed that intraCF contributes more to the phenomenon of forgetting in SFL than the model drifting of part-1, with the last layers playing an important role. Methods in related work concerning CL or  FL often utilize multiple blocks of higher layers, 
where each one of them is allocated to a different task,  
e.g., \citep{gurbuz2022nispa, kim2022multi, chen2023flexibility, hemati2025continual}. 
In such methods, often called \emph{multi-head}, the lower layers are shared among all tasks, 
while there exist multiple higher layers. We notice that the design of SFL naturally aligns with the multi-head architecture, as the higher layers are handled only by the server. Leveraging this together with the benefit of averaging under heterogeneous data~(as shown by the accuracy and PG  of FL in Fig.~\ref{fig:per_position_main}), we propose the \emph{first multi-head solution for SFL}, named~\textit{Hydra}, to tackle CF in SFL. Hydra derives from our insights in Sec.~\ref{sec:intraCF}.

\begin{figure}[t]
    \centering
\includegraphics[clip, width=0.99\columnwidth, trim={1.4cm 19.32cm 3.1cm 1.75cm}]{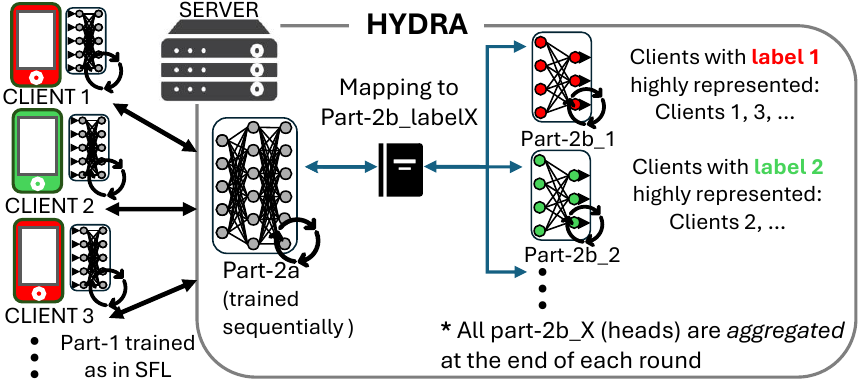}
\caption{The workflow of Hydra, the proposed method to mitigate catastrophic forgetting in SFL.
}
\label{fig:medusa_workflow}
\end{figure}

\noindent \textbf{Design Details.} Hydra extends SFL by modifying its workflow at the level of the server. Fig.~\ref{fig:medusa_workflow}  presents  Hydra’s design and workflow. In detail, part-1 is trained in the same way as before~(solely handled by the clients), while part-2~(managed by the server) is split and composed of: \textit{(i)}~a unique part-2a, shared among all clients and updated sequentially, and \textit{(ii)}~$G$ multiple versions of part-2b~(heads), which are updated in parallel and aggregated~(through \texttt{FedAvg}) at the end of each round.  Each head corresponds to a label or a superclass~(i.e., multiple labels). Further, the data corresponding to a group of clients is assigned to a specific head based on the highly represented labels in their data. This is done through a \emph{mapping} that the server performs, where it assigns the different outputs (activations) of part-2a to the different heads during forward propagation. A natural choice for the number of heads is the number of labels, i.e., $G = L$, following the multi-head approaches in CL. However, $G$ can be tuned according to the server's memory capacity to reduce overhead, as we discuss below.

The group of clients corresponding to each head is defined at the beginning of the training to cluster clients with similar data distributions, reducing large gradients that could overwrite previous updates. 
Homogeneity within a group ensures minimal weight changes,  
effectively addressing CF. This grouping requires each client to send information on the distribution of labels in its local data %(in a form of a vector of length $L$) 
(as a vector of length $L$) before the training starts. The \emph{grouping algorithm} is based on resource allocation problems and aims at grouping clients with similar label distributions while ensuring that the repartition of clients among the groups is balanced across all groups. In detail, it %decides the assignments of 
assigns clients to groups in a greedy way: for each group, the algorithm finds the client with the largest number of samples from the corresponding label and repeats this process until all clients have been assigned to exactly one group.
Finally, by iterating through the groups instead of the clients, the repartition of clients into the groups is balanced.
Note that the same mapping is then used inversely by the server during backward propagation. 

Unlike the existing multi-head approaches in CL, whose final trained model contains multiple heads, e.g., \cite{kim2022multi}, Hydra's final model contains a single head derived from aggregating all the different heads, making multi-head more suitable for SFL. %This is a key design feature that renders the multi-head approach more suitable for SFL. 
In fact, our evaluation~(Sec.~\ref{sec:exp}) shows that aggregated heads excel in performance when compared to typical multi-head architectures. To the best of our knowledge, Hydra is the first attempt to fine-tune the multi-head technique for SFL. Also,  other important design decisions of Hydra include the processing order of clients' activations through part-2a and the length of heads~(part-2b). To this end, we thoroughly study the impact of these design choices in Sec.~\ref{sec:exp}.  Finally, we note that Hydra differentiates from the  SL variant USplit~\cite{vepakomma2018split}, in which the model is split into 3 parts, and the last (3rd) model part is processed at the clients and aggregated~(at the end of each round). 
Indeed, in Hydra,  the number $G$ of heads is bounded by $L$,  making it less prone to model drifting w.r.t. USplit, where $C$ model parts are aggregated.

\begin{figure*}[!t]
    \centering
\includegraphics[width=2\columnwidth]{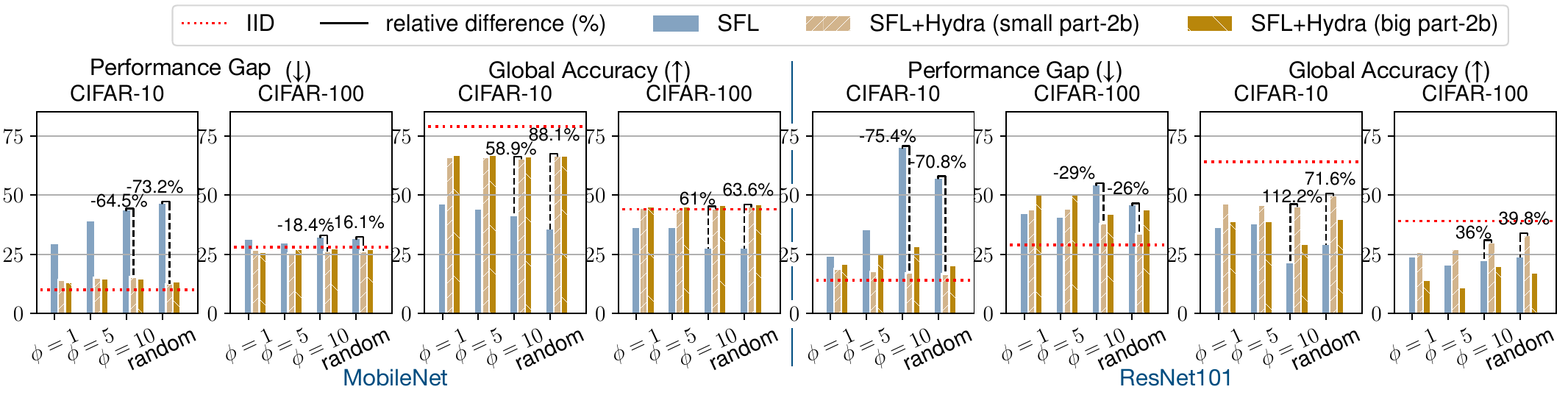}
\caption{Performance gap and global accuracy~(median) in SFL with and without Hydra with $80\%$-DL. Over-the-bar text shows the percentage of decrease/increase of the PG and global accuracy, respectively.
}
\label{fig:medusa_res}
\end{figure*}

\begin{table*}[!t]  
\centering 
\begin{tabular}{p{1.9cm}p{1.8cm}|p{1.2cm}p{1.8cm}p{1.8cm}|p{1.2cm}p{1.8cm}p{1.8cm}}
  & & \multicolumn{3}{c|}{\textbf{Global Accuracy $\uparrow$}} & \multicolumn{3}{c}{\textbf{Performance Gap $\downarrow$}}\\
 Dataset & Policy & Sharding &  Dirichlet \small{($\alpha=0.3$)} &  
 Dirichlet \small{($\alpha=0.1$)} &  Sharding &  Dirichlet \small{($\alpha=0.3$)} &   Dirichlet \small{($\alpha=0.1$)} \\ \hline 
\multirow[c]{1}{*}{CIFAR10} &  SFL+Hydra 
& $\mathbf{72}$ \small{$\mathbf{\pm 0.4}$} &  $\mathbf{59}$ \small{$\mathbf{\pm 0.4}$} &  $\mathbf{49}$ \small{$\mathbf{\pm 0.2}$}
& $\mathbf{10}$ \small{$\mathbf{\pm 0.4}$} &  $\mathbf{21}$ \small{$\mathbf{\pm 1}$} & $\mathbf{32}$ \small{$\mathbf{\pm 0.8}$} \\

\small{IID:(80, 10)} &  SFL 
& $50$ \small{$\pm 0.4$} &  $50$ \small{$\pm 0.5$} & $40$ \small{$\pm 0.6$}
&  $36$ \small{$\pm 1$} & $43.7$ \small{$\pm 0.5$} & $45.5$ \small{$\pm 1$} \\
\hline
\multirow[c]{1}{*}{TinyImageNet} & SFL+Hydra 
 & $\mathbf{34}$ \small{$\mathbf{\pm 0.1}$} & $\mathbf{45}$ \small{$\mathbf{\pm 0.1}$} &  $\mathbf{41}$ \small{$\mathbf{\pm 0.3}$}
 & $\mathbf{49}$ \small{$\mathbf{\pm 0.6}$} & $\mathbf{44}$ \small{$\mathbf{\pm 0.6}$} & $\mathbf{44.5}$ \small{$\mathbf{\pm 1}$}
 \\

\small{IID:(50, 40)}   &  SFL  
 & $27$ \small{$\pm 0.3$} & $\mathbf{45}$ \small{$\mathbf{\pm 0.3}$} &  $31$ \small{$\pm 1$}
 & $62$ \small{$\pm 1.6$} & $46$ \small{$\pm 1$} & $63$ \small{$\pm 0.7$} \\
\hline
\end{tabular}

\caption{Global accuracy and performance gap achieved by MobileNet with $\phi=10$ under different data partitions.}
\label{tab:other_data_pattern}
\end{table*}

\noindent \textbf{Hydra's Overhead.}  In terms of computing cost, there is a one-time $O(G \cdot C)$ cost for running the grouping algorithm at the beginning of training to assign clients into the different heads. During the training of part-2, there is no additional overhead since only one head is updated per batch update. 
Regarding the aggregation of heads at the end of every round, this can be implemented locally at the server and run in parallel with the aggregation of part-1. We stress that Hydra induces 
no computational overhead on the clients' side, since it only concerns part-2 of the model~(at the server). 

In terms of memory overhead, this depends mainly on the number $G$  and the length of the heads, but also on the implementation specifics. In practice, when the heads consist of only the last 2 layers, \emph{the memory overhead can be as little as} $G\cdot 6$MB, or $G\cdot10$MB when the heads consist of the last 6 layers. This estimation concerns Mobilenet and CIFAR-10 with a batch size of $64$, and is based on the memory needed for storing the model weights and the intermediate activations/gradients. In fact, our evaluation in Sec.~\ref{sec:exp} encourages small heads since Hydra performs better when compared to the case with larger heads.  Similarly, we show that Hydra leads to reduced CF when compared to vanilla SFL, even with a small $G$.
Finally, Hydra induces a negligible communication overhead at the beginning of the training, where each client sends a vector containing the information on its label distribution~(that is needed for the mapping algorithm).

\section{Experiments}
\label{sec:exp}

In this section, we thoroughly evaluate the performance of the proposed method Hydra. Fig.~\ref{fig:medusa_res} presents the results from SFL with and without Hydra~(denoted by SFL+Hydra and SFL, respectively). 
These results concern  multiple aspects:
the processing order of part-2a at the server, datasets, ML
models, and the selection of the heads' length.

\noindent \textbf{Processing Order of Part-2a.}  
We first focus on the case where  Hydra's heads  are small, and, in particular, 
their length is only 2 layers. This means that the second cut layer (i.e., the last layer of part-2a) is located at layer $24$ and $33$ for MobileNet and ResNet101, respectively. 
We notice that Hydra (SFL+Hydra) under \textbf{cyclic order} (with any $\phi$) significantly outperforms the baseline (SFL).
In detail, by employing Hydra, the labels' performance gap decreases by up to $64.5\%$ and $75.4\%$ for MobileNet and ResNet101 on CIFAR-10, respectively, while, at the same time,  the global accuracy increases by $58.9\%$ and $112.2\%$, respectively, approaching the accuracy under IID data.  Furthermore, Hydra has a robust performance regardless of the client scale~($\phi$), unlike the baseline SFL. From the results for the \textbf{random order}, we observe that Hydra reduces PG by up to $73\%$, thus providing a stable training even under a random order.

\noindent \textbf{Length of Heads (Part-2b) in Hydra.} 
We focus now (in Fig.~\ref{fig:medusa_res}) on Hydra with small and large lengths of heads~(e.g., for the large heads, the second cut layer is at layers $20$ and $30$ for MobileNet and ResNet101, respectively).
In general, for MobileNet, the two cases perform similarly, i.e,  Hydra with small heads achieves slightly better PG, while Hydra with larger heads has slightly better accuracy, indicating that MobileNet is not affected by the heads' length. However, ResNet101,  a more complex model, seems to be more sensitive to the length of heads since Hydra with small heads always outperforms the case of larger heads,  in terms of both  accuracy and PG, a side effect of model drifting. Therefore, a small length of heads seems to be the best design choice not only in terms of  memory overhead~(as discussed in Sec.~\ref{sec:medusa}), but also in terms of performance.

\noindent \textbf{Types of Data Partitioning.}  In Sec.~\ref{sec:medusa}, Hydra was defined for any data partitioning. While Fig.~\ref{fig:medusa_res} presented evaluation results of Hydra for the DL partition, Table~\ref{tab:other_data_pattern} presents   results  in three additional types of data partitioning, namely Dirichlet with $\alpha =0.1$ and $0.3$~(where lower $\alpha$ indicates higher heterogeneity),
and Sharding method with two dominant labels. The results demonstrate that \textit{Hydra improves over the baseline (SFL) in all the tested data partitioning methods}.

\begin{table}[!t]  
\centering 
\begin{tabular}{p{2.2cm}|p{1.5cm}p{1.3cm}p{1.3cm}}

\textbf{Policy} & \textbf{Gl. Acc. $\uparrow$} & \textbf{PG} $\downarrow$ & \textbf{BW} $\downarrow$ \\ \hline

 SFL+Hydra & $\mathbf{66.5}$\small{$\mathbf{\pm 0.1}$} & $\mathbf{13.4}$\small{$\mathbf{\pm 1}$} &  $\mathbf{9}$\small{$\mathbf{\pm 0.9}$}\\  SplitFedV1/FL
 %~\citep{thapa2022splitfed}} 
 & $55$\small{$\pm 0.2$} & $15.7$\small{$\pm 1.5$} & $14$\small{$\pm 2.7$} \\ SplitFedV3
 %~\citep{madaan2022vulnerability}} 
 & $43$\small{$\pm 0.3$} & $22.6$\small{$\pm 0.8$}  &  $20$\small{$\pm 2.5$} \\  MultiHead
 %~\citep{chen2023flexibility} } 
 & $47$\small{$\pm 0.3$} & $36.6$\small{$\pm 0.9$}  & $25$\small{$\pm 2.6$} \\ MergeSFL
 %~\citep{liao2023mergesfl}}
 & $65$\small{$\pm 0.5$} & $21$\small{$\pm 1.8$} & $13$\small{$\pm 1.3$}\\ 
 SplitNN
 %~\citep{vepakomma2018split}} 
 & $53$\small{$\pm 1$} & $44$\small{$\pm 1$} & $11$\small{$\pm 0.3$}\\
 \hline 

\end{tabular}
\caption{Comparison of Hydra with state-of-the-art methods in MobileNet with CIFAR-10 and $80\%$-DL.}
\label{tab:baseline_mobilenet}
\end{table}

\begin{figure}
    \centering
    \includegraphics[width=0.93\linewidth]{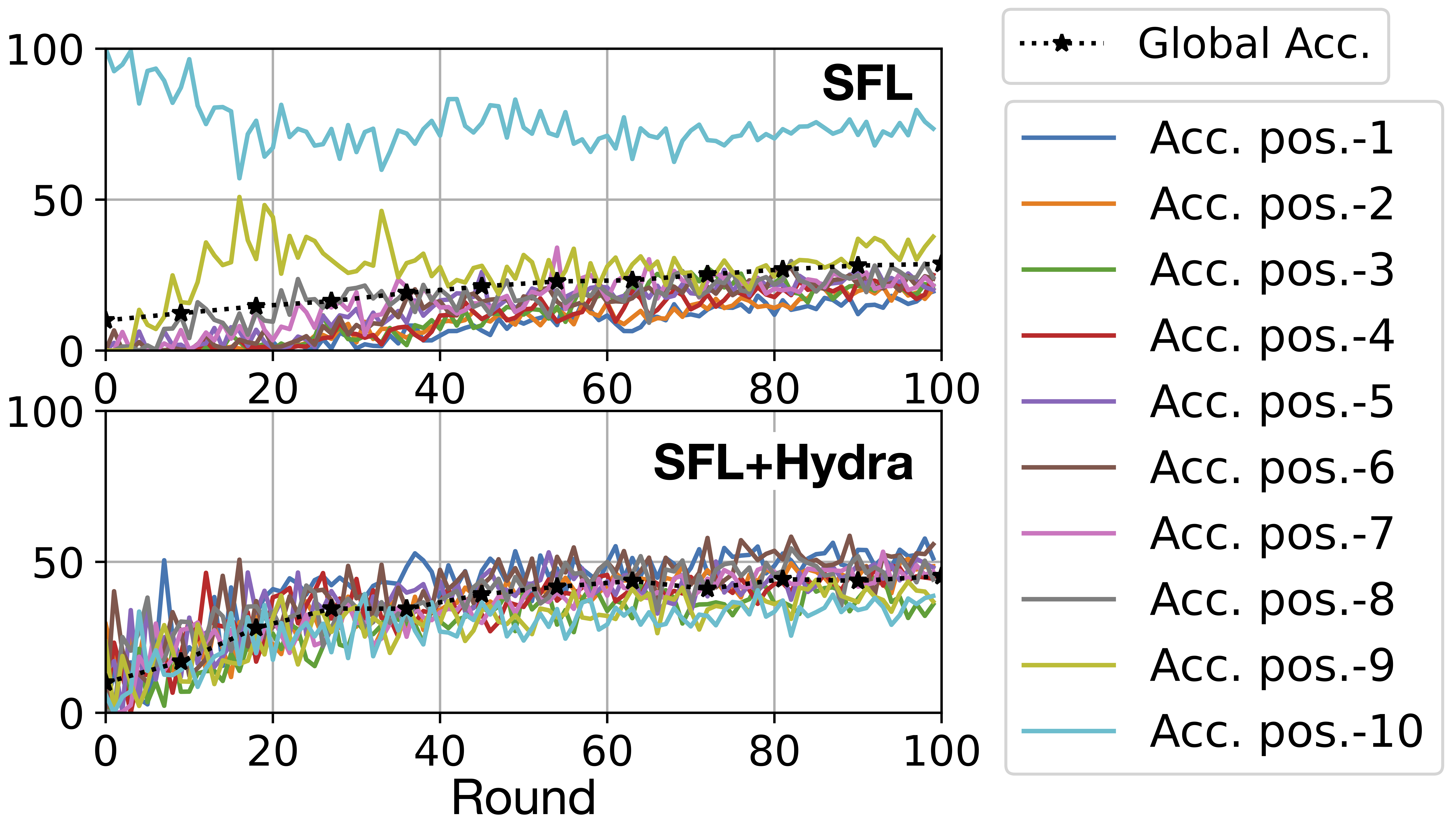}
    \caption{Per-position accuracy in SFL, and SFL+Hydra for ResNet101 with CIFAR-10 and $80\%$-DL partition.}
    \label{fig:per_pos_hydra_main}
\end{figure}

\begin{table}[t]  
\centering 
\begin{tabular}{p{1.36cm}p{0.6cm}|p{0.8cm}p{1.1cm}|p{1cm}p{1cm}}

 \textbf{Policy}& \textbf{N${^\circ}$} & \multicolumn{2}{c|}{\textbf{Gl. Accuracy $\uparrow$}} & \multicolumn{2}{c}{\textbf{Perf. Gap $\downarrow$}} \\ %\cline{1-1}
 & {\small\textbf{heads}} &  \textbf{cyclic} & \textbf{random} & \textbf{cyclic} & \textbf{random}\\ \hline
%\cline{1-6}
%\scriptsize{IID:(10, 80)}
 SFL   & N/A & $41.1$\scriptsize{$\pm 1$} & $35$\scriptsize{$\pm 4$} & $43.4 $\scriptsize{$\pm 3.6$} & $46.4$\scriptsize{$\pm 3$} \\
 SFL+Hydra & $10$ & 
 $\mathbf{65}$\scriptsize{$\mathbf{\pm 0.1}$}	& $\mathbf{66.5}$\scriptsize{$\mathbf{\pm 0.1}$} &
 $\mathbf{15.4}$\scriptsize{$\mathbf{\pm 0.6}$} & $\mathbf{13.4}$\scriptsize{$\mathbf{\pm 1}$}  \\
 SFL+Hydra & $5$ 
 & $56$\scriptsize{$\pm 1$} & $57$\scriptsize{$\pm 1$}
 & $19.3$\scriptsize{$\pm 1$} &  $23.1$\scriptsize{$\pm 1.5$}  \\
 SFL+Hydra & $2$ 
 & $49$\scriptsize{$\pm 1$} & $47$\scriptsize{$\pm 1$}
 & $32$\scriptsize{$\pm 2$} & $34$\scriptsize{$\pm 2$} \\ 
\hline
\end{tabular}
\caption{Ablation study  on the number of Hydra's heads~($G$) with MobileNet, CIFAR-10, $80\%$-DL, and IID:(80, 10).}
\label{tab:heads_ablation}
\end{table}

\noindent \textbf{Comparison with State-of-the-Art Methods.} 
In Table~\ref{tab:baseline_mobilenet},
we evaluate Hydra~(i.e., SFL+Hydra) using a random processing order~(i.e., capturing a more generic scenario) and MobileNet. We compare it with other policies, i.e., SplitFedV1/FL~\citep{thapa2022splitfed}, SplitFedV3~\citep{madaan2022vulnerability}, Multihead~\citep{chen2023flexibility},  MergeSFL~\citep{liao2023mergesfl}, and SplitNN~\cite{vepakomma2018split}.
All of these methods have different workflows  than SFL. Table~\ref{tab:baseline_mobilenet} shows that Hydra significantly outperforms the state-of-the-art. 
Specifically,   methods in the literature fail to improve on all metrics~(indicating an unstable training), while  \emph{Hydra consistently excels in all metrics~(accuracy, PG, and BW)}. Importantly,  Hydra outperforms the other multi-head approaches~(i.e., SplitFedV3 and MultiHead). This highlights that Hydra's aggregation of the heads is a key feature over the classic multi-head approaches. Moreover, MergeSFL achieves a global accuracy comparable to the one of Hydra. However, a significant drawback of MergeSFL lies in its implementation, which requires synchronization among clients and more frequent updates~(due to micro-batches). 
Finally, among all the methods in the literature, SplitNN is the only one where the server processes clients' data in a specific order~(typically random). As a result,  SplitNN has the worst~(i.e., highest) performance gap among all methods due to forgetting.

\noindent \textbf{Closing the Labels' Performance Gap.} To better understand Hydra's performance, we dive into the per-position accuracy~(i.e., the accuracy of the labels that
are dominant in clients that are processed at a certain position, see Sec.~\ref{sec:method_exper}). In Fig.~\ref{fig:per_pos_hydra_main}, we see that Hydra effectively reduces the labels' performance gap  and stabilizes the training, even in a highly heterogeneous scenario, i.e., $80\%$-DL. In detail, unlike the baseline~(SFL), there is no significant gap among the different positions. We note that the peaks and troughs in the subplot on the right are a result of the processing order under non-IID data. Finally, the global accuracy improves over SFL, i.e., $44\%$ instead of $28\%$, after 100 rounds.

\noindent \textbf{Ablation Study on the Number of Heads $\mathbf{G}$.} As discussed in Sec.~\ref{sec:medusa}, Hydra's memory overhead can be adjusted by reducing the length of part-2b and/or $G$. For the former, we showed that a smaller part-2b performs better.  Table~\ref{tab:heads_ablation} reports the performance of Hydra for different values of $G$ for CIFAR-10 and different processing orders of part-2a~(cyclic with $\phi=10$ and random). For CIFAR-10, we define $G$ superclasses based on the semantic relationships among labels, similar to~\citep{bai2021clustering}. For example, when $G=2$, we consider the two superclasses: animals and no-animals.
We see that Hydra performs best when $G=L$, but still outperforms the baseline SFL in all metrics, even for a smaller number of heads. 
These results highlight an interesting trade-off between memory overhead and performance.

\section{Conclusion and Discussion}
\label{sec:conclusions}

In this work, we studied the phenomenon of catastrophic forgetting as a result of data heterogeneity in SFL. Our empirical analysis was performed in a variety of scenarios~(e.g., ML models, datasets, etc.)
allowing us to identify key parameters and derive insights into CF in SFL. Based on these, we proposed Hydra, a novel mitigation method. We showed that it successfully alleviates the forgetting while increasing the global accuracy and closing the labels' performance gaps. Hydra induces minimal computing and memory overhead (that can be further reduced through design choices) and outperforms baselines and state-of-the-art methods.

\noindent \textbf{Limitations and Future Work.} Despite the absence of theoretical claims in this work, our empirical analysis of CF in SFL paves the way towards a theoretical/convergence analysis of SFL with respect to key parameters such as the cut layer and the processing order at the server. Such an analysis could also be the stepping stone towards establishing theoretical guarantees for Hydra. Furthermore, an interesting direction for future work is to incorporate client selection policies into the analysis of CF, but also in Hydra, similar to related work in FL, e.g.,~\citep{lee2022preservation}. Finally, our ablation study on the number of Hydra's heads revealed that semantic relationships among labels may play an important role in CF and in SFL's training in general. While this aspect was not the primary focus of this work, we believe that it should be studied in the future, as it has been done for continual learning~\citep{ramasesh2020anatomy}.

\section*{Acknowledgments}
This work was funded by the Horizon MSCA
Postdoctoral Fellowship OPALS (grant agreement 101210495), the Spanish Ministry of Economic Affairs
and Digital Transformation, the European Union-NextGenerationEU through the projects 6G-RIEMANN (TSI-063000-2021-147) and MAP-6G (TSI-063000-2021-63), and the Smart Networks and Services Joint Undertaking (SNS JU) under the European Union’s Horizon Europe research and innovation programme through the CONFIDENTIAL6G EU project (Grant Agreement No 101096435). Views and opinions expressed are however those of the author(s) only and do not necessarily reflect those of the European Union. Neither the European Union nor the granting authority can be held responsible for them.
\begin{center}
\includegraphics[width=3cm]{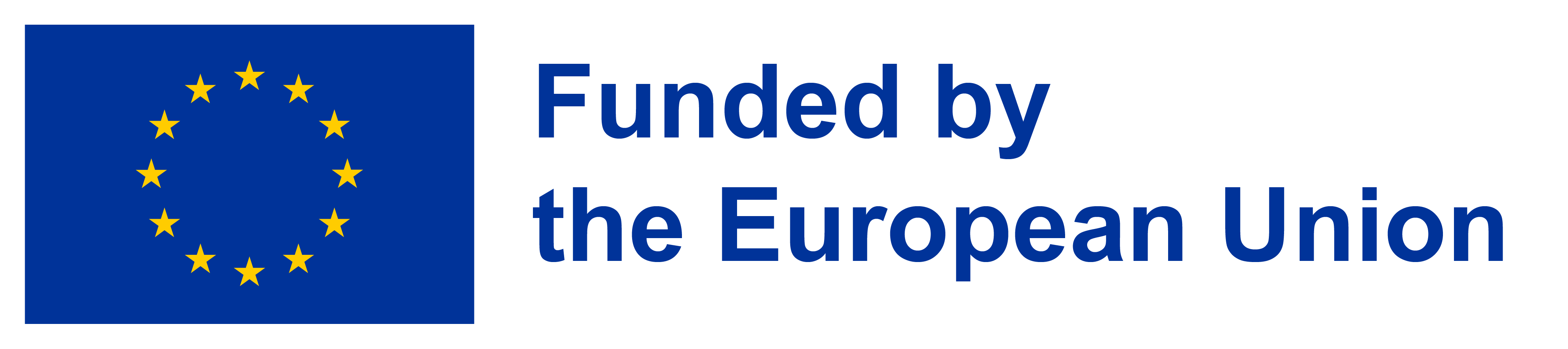}
\end{center}

\clearpage

\bibliographystyle{apalike}
\bibliography{aaai2026}

@article{feng2023iotsl,
  title={IoTSL: Toward Efficient Distributed Learning for Resource-Constrained Internet of Things},
  author={Feng, Xingyu and Luo, Chengwen and Chen, Jiongzhang and Huang, Yijing and Zhang, Jin and Xu, Weitao and Li, Jianqiang and Leung, Victor CM},
  journal={IEEE Internet of Things Journal},
  volume={10},
  number={11},
  pages={9892--9905},
  year={2023},
  publisher={IEEE}
}

@incollection{madaan2022vulnerability,
  title={Vulnerability due to training order in split learning},
  author={Madaan, Harshit and Kulkarni, Manish Gawali Viraj and Pant, Aniruddha},
  booktitle={ICT Systems and Sustainability: Proceedings of ICT4SD 2021, Volume 1},
  pages={103--112},
  year={2022},
  publisher={Springer}
}

@inproceedings{xia2024korea,
  title={MultiSFL: Towards Accurate Split Federated Learning via Multi-Model Aggregation and Knowledge Replay},
  author={Xia, Zeke and Hu, Ming and Yan, Dengke and Liu, Ruixuan and Li, Anran and Xie, Xiaofei and Chen, Mingsong},
  booktitle={Proceedings of the AAAI Conference on Artificial Intelligence},
  volume={39},
  number={1},
  pages={914--922},
  year={2025}
}

@article{goodfellow2013empirical,
  title={An empirical investigation of catastrophic forgetting in gradient-based neural networks},
  author={Goodfellow, Ian J and Mirza, Mehdi and Xiao, Da and Courville, Aaron and Bengio, Yoshua},
  journal={arXiv preprint arXiv:1312.6211},
  year={2013}
}

@article{wang2024comprehensive,
  title={A comprehensive survey of continual learning: theory, method and application},
  author={Wang, Liyuan and Zhang, Xingxing and Su, Hang and Zhu, Jun},
  journal={IEEE Transactions on Pattern Analysis and Machine Intelligence},
  year={2024},
  publisher={IEEE}
}

@inproceedings{ma2022continual,
  title={Continual Federated Learning Based on Knowledge Distillation.},
  author={Ma, Yuhang and Xie, Zhongle and Wang, Jue and Chen, Ke and Shou, Lidan},
  booktitle={IJCAI},
  pages={2182--2188},
  year={2022}
}

@article{kirkpatrick2017overcoming,
  title={Overcoming catastrophic forgetting in neural networks},
  author={Kirkpatrick, James and Pascanu, Razvan and Rabinowitz, Neil and Veness, Joel and Desjardins, Guillaume and Rusu, Andrei A and Milan, Kieran and Quan, John and Ramalho, Tiago and Grabska-Barwinska, Agnieszka and others},
  journal={Proceedings of the national academy of sciences},
  volume={114},
  number={13},
  pages={3521--3526},
  year={2017},
  publisher={National Acad Sciences}
}

@article{xu2022acceleration,
  title={Acceleration of federated learning with alleviated forgetting in local training},
  author={Xu, Chencheng and Hong, Zhiwei and Huang, Minlie and Jiang, Tao},
  journal={arXiv preprint arXiv:2203.02645},
  year={2022}
}

@inproceedings{
ramasesh2020anatomy,
title={Anatomy of Catastrophic Forgetting: Hidden Representations and Task Semantics},
author={Vinay Venkatesh Ramasesh and Ethan Dyer and Maithra Raghu},
booktitle={International Conference on Learning Representations},
year={2021}
}

@inproceedings{karimireddy2020scaffold,
  title={Scaffold: Stochastic controlled averaging for federated learning},
  author={Karimireddy, Sai Praneeth and Kale, Satyen and Mohri, Mehryar and Reddi, Sashank and Stich, Sebastian and Suresh, Ananda Theertha},
  booktitle={International conference on machine learning},
  pages={5132--5143},
  year={2020},
  organization={PMLR}
}

@inproceedings{thapa2022splitfed,
  title={Splitfed: When federated learning meets split learning},
  author={Thapa, Chandra and Arachchige, Pathum Chamikara Mahawaga and Camtepe, Seyit and Sun, Lichao},
  booktitle={Proceedings of the AAAI Conference on Artificial Intelligence},
  volume={36},
  pages={8485--8493},
  year={2022}
}

@article{hafi2024split,
  title={Split federated learning for 6G enabled-networks: Requirements, challenges and future directions},
  author={Hafi, Houda and Brik, Bouziane and Frangoudis, Pantelis A and Ksentini, Adlen and Bagaa, Miloud},
  journal={IEEE Access},
  volume={12},
  pages={9890--9930},
  year={2024},
  publisher={IEEE}
}

@inproceedings{tirana2024workflow,
  title={Workflow Optimization for Parallel Split Learning},
  author={Tirana, Joana and Tsigkari, Dimitra and Iosifidis, George and Chatzopoulos, Dimitris},
  booktitle={IEEE INFOCOM 2024 - IEEE Conference on Computer Communications}, 
  pages={1331-1340},
  year={2024}
}

@inproceedings{liao2023mergesfl,
  title={Mergesfl: Split federated learning with feature merging and batch size regulation},
  author={Liao, Yunming and Xu, Yang and Xu, Hongli and Wang, Lun and Yao, Zhiwei and Qiao, Chunming},
  booktitle={2024 IEEE 40th International Conference on Data Engineering (ICDE)},
  pages={2054--2067},
  year={2024},
  organization={IEEE}
}

@article{zhu2021federated,
  title={Federated learning on non-IID data: A survey},
  author={Zhu, Hangyu and Xu, Jinjin and Liu, Shiqing and Jin, Yaochu},
  journal={Neurocomputing},
  volume={465},
  pages={371--390},
  year={2021},
  publisher={Elsevier}
}

@article{liu2022wireless,
  title={Wireless distributed learning: A new hybrid split and federated learning approach},
  author={Liu, Xiaolan and Deng, Yansha and Mahmoodi, Toktam},
  journal={IEEE Transactions on Wireless Communications},
  volume={22},
  number={4},
  pages={2650--2665},
  year={2022},
  publisher={IEEE}
}

@inproceedings{ramasesh2021effect,
  title={Effect of scale on catastrophic forgetting in neural networks},
  author={Ramasesh, Vinay Venkatesh and Lewkowycz, Aitor and Dyer, Ethan},
  booktitle={International Conference on Learning Representations},
  year={2021}
}

@inproceedings{lin2023theory,
  title={Theory on forgetting and generalization of continual learning},
  author={Lin, Sen and Ju, Peizhong and Liang, Yingbin and Shroff, Ness},
  booktitle={International Conference on Machine Learning},
  pages={21078--21100},
  year={2023},
  organization={PMLR}
}

@article{lee2022preservation,
  title={Preservation of the global knowledge by not-true distillation in federated learning},
  author={Lee, Gihun and Jeong, Minchan and Shin, Yongjin and Bae, Sangmin and Yun, Se-Young},
  journal={Advances in Neural Information Processing Systems},
  volume={35},
  pages={38461--38474},
  year={2022}
}

@article{li2020federated,
  title={Federated optimization in heterogeneous networks},
  author={Li, Tian and Sahu, Anit Kumar and Zaheer, Manzil and Sanjabi, Maziar and Talwalkar, Ameet and Smith, Virginia},
  journal={Proceedings of Machine learning and systems},
  volume={2},
  pages={429--450},
  year={2020}
}

@article{wang2020tackling,
  title={Tackling the objective inconsistency problem in heterogeneous federated optimization},
  author={Wang, Jianyu and Liu, Qinghua and Liang, Hao and Joshi, Gauri and Poor, H Vincent},
  journal={Advances in neural information processing systems},
  volume={33},
  pages={7611--7623},
  year={2020}
}

@article{swartworth2023nearly,
  title={Nearly optimal bounds for cyclic forgetting},
  author={Swartworth, William and Needell, Deanna and Ward, Rachel and Kong, Mark and Jeong, Halyun},
  journal={Advances in Neural Information Processing Systems},
  volume={36},
  pages={68197--68206},
  year={2023}
}

@article{hemati2025continual,
  title={Continual learning in the presence of repetition},
  author={Hemati, Hamed and Pellegrini, Lorenzo and Duan, Xiaotian and Zhao, Zixuan and Xia, Fangfang and Masana, Marc and Tscheschner, Benedikt and Veas, Eduardo and Zheng, Yuxiang and Zhao, Shiji and others},
  journal={Neural Networks},
  volume={183},
  pages={106920},
  year={2025},
  publisher={Elsevier}
}

@inproceedings{li2022federated,
  title={Federated learning on non-iid data silos: An experimental study},
  author={Li, Qinbin and Diao, Yiqun and Chen, Quan and He, Bingsheng},
  booktitle={2022 IEEE 38th International Conference on Data Engineering (ICDE)},
  pages={965--978},
  year={2022},
  organization={IEEE}
}

@article{kairouz2021advances,
  title={Advances and open problems in federated learning},
  author={Kairouz, Peter and McMahan, H Brendan and Avent, Brendan and Bellet, Aur{\'e}lien and Bennis, Mehdi and Bhagoji, Arjun Nitin and Bonawitz, Kallista and Charles, Zachary and Cormode, Graham and Cummings, Rachel and others},
  journal={Foundations and Trends{\textregistered} in Machine Learning},
  volume={14},
  number={1--2},
  pages={1--210},
  year={2021},
  publisher={Now Publishers, Inc.}
}

@article{kang2017neurosurgeon,
  title={Neurosurgeon: Collaborative intelligence between the cloud and mobile edge},
  author={Kang, Yiping and Hauswald, Johann and Gao, Cao and Rovinski, Austin and Mudge, Trevor and Mars, Jason and Tang, Lingjia},
  journal={ACM SIGARCH Computer Architecture News},
  volume={45},
  number={1},
  pages={615--629},
  year={2017},
  publisher={ACM New York, NY, USA}
}

@article{solans2024non,
  title={Non-IID data in Federated Learning: A Systematic Review with Taxonomy, Metrics, Methods, Frameworks and Future Directions},
  author={Jimenez G., Daniel M. and Solans, David and Heikkila, Mikko and Vitaletti, Andrea and Kourtellis, Nicolas and Anagnostopoulos, Aris and Chatzigiannakis, Ioannis and others},
  journal={arXiv preprint arXiv:2411.12377},
  year={2024}
}

@inproceedings{gurbuz2022nispa,
  title={NISPA: Neuro-Inspired Stability-Plasticity Adaptation for Continual Learning in Sparse Networks},
  author={Gurbuz, Mustafa B and Dovrolis, Constantine},
  booktitle={International Conference on Machine Learning},
  pages={8157--8174},
  year={2022},
  organization={PMLR}
}

@inproceedings{mcmahan2017communication,
  title={Communication-efficient learning of deep networks from decentralized data},
  author={McMahan, Brendan and Moore, Eider and Ramage, Daniel and Hampson, Seth and y Arcas, Blaise Aguera},
  booktitle={Artificial intelligence and statistics},
  pages={1273--1282},
  year={2017},
  organization={PMLR}
}

@article{zilly2021plasticity,
  title={On plasticity, invariance, and mutually frozen weights in sequential task learning},
  author={Zilly, Julian and Achille, Alessandro and Censi, Andrea and Frazzoli, Emilio},
  journal={Advances in Neural Information Processing Systems},
  volume={34},
  pages={12386--12399},
  year={2021}
}

@article{krizhevsky2012imagenet,
  title={Imagenet classification with deep convolutional neural networks},
  author={Krizhevsky, Alex and Sutskever, Ilya and Hinton, Geoffrey E},
  journal={Advances in neural information processing systems},
  volume={25},
  year={2012}
}

@inproceedings{he2016deep,
  title={Deep residual learning for image recognition},
  author={He, Kaiming and Zhang, Xiangyu and Ren, Shaoqing and Sun, Jian},
  booktitle={Proc. of the IEEE conference on computer vision and pattern recognition},
  pages={770--778},
  year={2016}
}

@article{simonyan2014very,
  title={Very deep convolutional networks for large-scale image recognition},
  author={Simonyan, Karen and Zisserman, Andrew},
  journal={arXiv preprint arXiv:1409.1556},
  year={2014}
}

@inproceedings{kim2022multi,
  title={A multi-head model for continual learning via out-of-distribution replay},
  author={Kim, Gyuhak and Liu, Bing and Ke, Zixuan},
  booktitle={Conference on Lifelong Learning Agents},
  pages={548--563},
  year={2022},
  organization={PMLR}
}

@inproceedings{chen2023flexibility,
  title={Flexibility and Privacy: A Multi-Head Federated Continual Learning Framework for Dynamic Edge Environments},
  author={Chen, Chunlu and Kevin, I and Wang, Kai and Li, Peng and Sakurai, Kouichi},
  booktitle={2023 Eleventh International Symposium on Computing and Networking (CANDAR)},
  pages={1--10},
  year={2023},
  organization={IEEE}
}

@article{feng2024slwf,
  title={{SLwF}: A Split Learning Without Forgetting Framework for Internet of Things},
  author={Feng, Xingyu and Jia, Renqi and Luo, Chengwen and Leung, Victor CM and Xu, Weitao},
  journal={IEEE Internet of Things Journal},
  year={2024},
  publisher={IEEE}
}

@inproceedings{hu2024fedmut,
  title={FedMut: Generalized federated learning via stochastic mutation},
  author={Hu, Ming and Cao, Yue and Li, Anran and Li, Zhiming and Liu, Chengwei and Li, Tianlin and Chen, Mingsong and Liu, Yang},
  booktitle={Proceedings of the AAAI conference on artificial intelligence},
  volume={38},
  number={11},
  pages={12528--12537},
  year={2024}
}

@article{zhao2018federated,
  title={Federated learning with non-iid data},
  author={Zhao, Yue and Li, Meng and Lai, Liangzhen and Suda, Naveen and Civin, Damon and Chandra, Vikas},
  journal={arXiv preprint arXiv:1806.00582},
  year={2018}
}

@article{vepakomma2018split,
  title={Split learning for health: Distributed deep learning without sharing raw patient data},
  author={Vepakomma, Praneeth and Gupta, Otkrist and Swedish, Tristan and Raskar, Ramesh},
  journal={arXiv preprint arXiv:1812.00564},
  year={2018}
}

@inproceedings{gawali2021comparison,
  title={Comparison of privacy-preserving distributed deep learning methods in healthcare},
  author={Gawali, Manish and Arvind, CS and Suryavanshi, Shriya and Madaan, Harshit and Gaikwad, Ashrika and Bhanu Prakash, KN and Kulkarni, Viraj and Pant, Aniruddha},
  booktitle={Medical Image Understanding and Analysis: 25th Annual Conference, MIUA},
  pages={457--471},
  year={2021},
  organization={Springer}
}

@article{howard2017mobilenets,
  title={MobileNets: Efficient Convolutional Neural Networks for Mobile Vision Applications. sl, sn},
  author={Howard, AG and Zhu, M and Chen, B and Kalenichenko, D and Wang, W and Weyand, T and Andreetto, M and Adam, H},
  journal={arXiv preprint arXiv:1704.04861},
  year={2017}
}

@inproceedings{gao2020end,
  title={End-to-End Evaluation of Federated Learning and Split Learning for Internet of Things},
  author={Gao, Yansong and Kim, Minki and Abuadbba, Sharif and Kim, Yeonjae and Thapa, Chandra and Kim, Kyuyeon and Camtep, Seyit A and Kim, Hyoungshick and Nepal, Surya},
  booktitle={2020 International Symposium on Reliable Distributed Systems (SRDS)},
  pages={91--100},
  year={2020},
  organization={IEEE Computer Society}
}

@inproceedings{jeon2020privacy,
  title={Privacy-sensitive parallel split learning},
  author={Jeon, Joohyung and Kim, Joongheon},
  booktitle={2020 International Conference on Information Networking (ICOIN)},
  pages={7--9},
  year={2020},
  organization={IEEE}
}

@misc{minka2000dirichlet,
  title={Estimating a Dirichlet distribution},
  author={Minka, Thomas},
  year={2000},
  publisher={Technical report, MIT}
}

@inproceedings{zhang2021federated,
  title={Federated learning for non-iid data via unified feature learning and optimization objective alignment},
  author={Zhang, Lin and Luo, Yong and Bai, Yan and Du, Bo and Duan, Ling-Yu},
  booktitle={Proceedings of the IEEE/CVF international conference on computer vision},
  pages={4420--4428},
  year={2021}
}

@inproceedings{kemker2018measuring,
  title={Measuring catastrophic forgetting in neural networks},
  author={Kemker, Ronald and McClure, Marc and Abitino, Angelina and Hayes, Tyler and Kanan, Christopher},
  booktitle={Proceedings of the AAAI conference on artificial intelligence},
  volume={32},
  number={1},
  year={2018}
}

@inproceedings{bengio2009curriculum,
  title={Curriculum learning},
  author={Bengio, Yoshua and Louradour, J{\'e}r{\^o}me and Collobert, Ronan and Weston, Jason},
  booktitle={Proceedings of the 26th annual international conference on machine learning},
  pages={41--48},
  year={2009}
}

@inproceedings{stojanov2019incremental,
  title={Incremental object learning from contiguous views},
  author={Stojanov, Stefan and Mishra, Samarth and Thai, Ngoc Anh and Dhanda, Nikhil and Humayun, Ahmad and Yu, Chen and Smith, Linda B and Rehg, James M},
  booktitle={Proceedings of the ieee/cvf conference on computer vision and pattern recognition},
  pages={8777--8786},
  year={2019}
}

@inproceedings{evron2023continual,
  title={Continual learning in linear classification on separable data},
  author={Evron, Itay and Moroshko, Edward and Buzaglo, Gon and Khriesh, Maroun and Marjieh, Badea and Srebro, Nathan and Soudry, Daniel},
  booktitle={International Conference on Machine Learning~(ICML)},
  pages={9440--9484},
  year={2023},
  organization={PMLR}
}

@article{cossu2022class,
  title={Is class-incremental enough for continual learning?},
  author={Cossu, Andrea and Graffieti, Gabriele and Pellegrini, Lorenzo and Maltoni, Davide and Bacciu, Davide and Carta, Antonio and Lomonaco, Vincenzo},
  journal={Frontiers in Artificial Intelligence},
  volume={5},
  pages={829842},
  year={2022},
  publisher={Frontiers Media SA}
}

@inproceedings{hanconvergence,
  title={Convergence Analysis of Split Federated Learning on Heterogeneous Data},
  author={Han, Pengchao and Huang, Chao and Tian, Geng and Tang, Ming and Liu, Xin},
  booktitle={The Thirty-eighth Annual Conference on Neural Information Processing Systems},
  year={2024}
}

@article{li2023convergence,
  title={Convergence analysis of sequential federated learning on heterogeneous data},
  author={Li, Yipeng and Lyu, Xinchen},
  journal={Advances in Neural Information Processing Systems},
  volume={36},
  pages={56700--56755},
  year={2023}
}

@inproceedings{netzer2011reading,
  title={Reading digits in natural images with unsupervised feature learning},
  author={Netzer, Yuval and Wang, Tao and Coates, Adam and Bissacco, Alessandro and Wu, Baolin and Ng, Andrew Y and others},
  booktitle={NIPS workshop on deep learning and unsupervised feature learning},
  volume={2011},
  number={2},
  pages={4},
  year={2011},
  organization={Granada}
}

@article{krizhevsky2009learning,
  title={Learning multiple layers of features from tiny images},
  author={Krizhevsky, Alex and Hinton, Geoffrey and others},
  year={2009},
  publisher={Toronto, ON, Canada}
}

@article{le2015tiny,
  title={Tiny imagenet visual recognition challenge},
  author={Le, Yann and Yang, Xuan},
  journal={CS 231N},
  volume={7},
  number={7},
  pages={3},
  year={2015}
}

@article{bai2021clustering,
  title={Clustering effect of adversarial robust models},
  author={Bai, Yang and Yan, Xin and Jiang, Yong and Xia, Shu-Tao and Wang, Yisen},
  journal={Advances in Neural Information Processing Systems},
  volume={34},
  pages={29590--29601},
  year={2021}
}

@article{leys2013detecting,
  title={Detecting outliers: Do not use standard deviation around the mean, use absolute deviation around the median},
  author={Leys, Christophe and Ley, Christophe and Klein, Olivier and Bernard, Philippe and Licata, Laurent},
  journal={Journal of experimental social psychology},
  volume={49},
  number={4},
  pages={764--766},
  year={2013},
  publisher={Elsevier}
}

@inproceedings{evron2022catastrophic,
  title={How catastrophic can catastrophic forgetting be in linear regression?},
  author={Evron, Itay and Moroshko, Edward and Ward, Rachel and Srebro, Nathan and Soudry, Daniel},
  booktitle={Conference on Learning Theory},
  pages={4028--4079},
  year={2022},
  organization={PMLR}
}

@inproceedings{li2024towards,
  title={Towards efficient replay in federated incremental learning},
  author={Li, Yichen and Li, Qunwei and Wang, Haozhao and Li, Ruixuan and Zhong, Wenliang and Zhang, Guannan},
  booktitle={Proceedings of the IEEE/CVF Conference on Computer Vision and Pattern Recognition},
  pages={12820--12829},
  year={2024}
}

@inproceedings{qibetter,
  title={Better Generative Replay for Continual Federated Learning},
  author={Qi, Daiqing and Zhao, Handong and Li, Sheng},
  booktitle={The Eleventh International Conference on Learning Representations},
  year={2023}
}

@article{gao2021evaluation,
  title={Evaluation and optimization of distributed machine learning techniques for internet of things},
  author={Gao, Yansong and Kim, Minki and Thapa, Chandra and Abuadbba, Alsharif and Zhang, Zhi and Camtepe, Seyit and Kim, Hyoungshick and Nepal, Surya},
  journal={IEEE Transactions on Computers},
  volume={71},
  number={10},
  pages={2538--2552},
  year={2021},
  publisher={IEEE}
}

\appendix
  \onecolumn
 
{\large \textbf{Overview of Appendix}}

\begin{itemize}
     \item \textbf{Appendix~\ref{ap:hydra}} discusses the design details of the proposed approach, \emph{Hydra}, i.e., the proposed algorithm and the grouping algorithm. Hydra is first presented in Section~\ref{sec:medusa}.
    \item \textbf{Appendix~\ref{ap:protocols}} presents the overview of several variants of Split Federated Learning~(SFL), and the multi-head Federated Learning~(FL) approach. These methods are used as baseline algorithms during the numerical evaluation.
    \item \textbf{Appendix~\ref{ap:ext_related_work}} extends the Related Work section~(Section~\ref{sec:related_work}) and contains background information regarding weight regularization methods used in Continual Learning~(CL). 
    \item \textbf{Appendix~\ref{ap:implem_details}} presents the details of the experimental set-up used throughout the paper.
    \item \textbf{Appendix~\ref{ap:sim_data_hetero}} elaborates on the types of data partitions used in this paper to simulate  data heterogeneity.
   
    \item \textbf{Appendix~\ref{ap:other_orders}} contains details for one of the processing orders studied in the paper~(cyclic order), as well as an additional one, the cyclic-and-reverse order.
    
    \item \textbf{Appendix~\ref{ap:ext_syst_analasis}} presents additional experiments of the systematic analysis of Catastrophic Forgetting~(CF) in Section~\ref{sec:intraCF}.
    \item \textbf{Appendix~\ref{ap:extend_evaluation}} provides additional results complementing the numerical evaluation of Hydra in
    Section~\ref{sec:exp}.
\end{itemize}

%\noindent \textbf{Note:} References to the main paper's Figures and Tables are differentiated with the \emph{-M} and italic font.

\normalsize

\section{Hydra Design Details}
\label{ap:hydra}

\subsection{\revision{Algorithm of Hydra}}
\label{ap:hydra_overview}

Algorithm~\ref{alg:hydra_alg} presents in detail the workflow of Hydra.\footnote{We note that part of this workflow is the same as (or aligns with) the workflow of SFL, e.g., how part-1 of the model is processed. For this reason, we often refer to it as SFL+Hydra.}
The algorithm describes the steps of the three main entities~(i.e., Aggregator, Server, and  Clients). In detail,  the aggregator initiates the training by initializing the weights of the model, and   sends the corresponding model parts to the server and the clients, i.e., weights $w_{p1}$ of part-1 are sent to clients, and weights $w_{p2_a}, w_{p2_b}$ of part-2a and part-2b, are sent to the server. Then the server and the client can start training.

At the beginning of the training, the group of clients corresponding to each head~(note that the server maintains $G$  heads) is defined to group clients with similar data distributions. This grouping requires each client to send information on the distribution of labels in its local data before the training starts.  The mapping (or grouping) of each client's intermediate data to the different part-2b heads is done by using the $u$ set, which contains the mapping of clients into groups. This set is defined by Algorithm~\ref{alg:client_assigment}, following the assignment policy described in Appendix~\ref{ap:group_def}. 

Next, the training takes place in $R$ rounds. In every round, the server  ensures that it receives the batch updates from all clients. Note that the server `pulls' a new batch request~(which contains the activation from a client) using the function \verb|proc_order()|. This function guarantees that the preferred processing order is followed, e.g., cyclic order (or cyclic-and-reverse, as in Appendix~\ref{ap:permutation}) or random order.

Finally, at the end of each round, the aggregation takes place. In detail, the clients send the updated weights of part-1 to the aggregator, which will compute the weights for the next round, using an aggregation function, like \texttt{FedAvg}. We note that Hydra's design provides the flexibility to apply any aggregation function, as it does not alter the processing workflow.  Concurrently, the server aggregates \textit{locally} all the different heads.

\subsection{The Grouping Algorithm of Hydra}
\label{ap:group_def}

In this section, we describe in detail how clients' data are mapped to different versions of part-2b (heads). In its core, this mapping \emph{groups} clients according to the similarity in their data. Therefore, we often use the terms  ``heads'' and ``groups'' interchangeably.  Let $g \in \mathcal{G}$ represent a group, where $\mathcal{G}$ is the set of all groups, and let  $G = |\mathcal{G}|$ denote the total number of groups~(each group corresponds to a  part-2b head, as depicted in Figure~\ref{fig:medusa_workflow}). Hydra's design naturally bounds $G$ by the number of different labels, i.e., $G\leq L$.  This is in line with the multi-head architectures in CL, where, typically, the number of heads is equal to the number of tasks.  In what follows, for simplicity, we describe the grouping algorithm of Hydra assuming that $G = L$. However, this description can be easily extended for the case where  $G < L$, as we do in our ablation study in Section~\ref{sec:exp}.

\begin{algorithm*}[h]
   \caption{Overview of SFL+Hydra}
   \label{alg:hydra_alg}
\begin{algorithmic}

    \STATE {\bfseries Function:}  AggregatorSide
        \STATE \hspace{0.4cm} $w^0_{p1};w^0_{p2_a};w^0_{p2_b}  \leftarrow$ init() \hspace{1.5cm}   \textit{// initialize model parts (part-1, part-2a, and part-2b)}
        \STATE \hspace{0.4cm} send $(w^0_{p2_a};w^0_{p2_b})$ to server
        \STATE \hspace{0.4cm} {\bfseries for} {$r=1$ {\bfseries to} $R$} 
            \STATE \hspace{0.8cm} send $(w^r_{p1})$ to all clients
            \STATE \hspace{0.8cm} wait until: receive   $(w^{rc}_{p1})$ from all clients \hspace{0.2cm}   \textit{// wait for round to complete and gather weight updates from all clients}
            \STATE \hspace{0.8cm} $w^{r+1}_{p1} \leftarrow$ aggr($\{w^{rc}_{p1} \textit{, } \forall c \in \mathcal{C}\})$ \hspace{1.5cm}   \textit{// aggregate all weights updates from clients}
        \STATE  \hspace{0.4cm} {\bfseries end for}
    \STATE {\bfseries EndFunction} 
    
    % \STATE 
    % \STATE 
    \vspace{0.2cm}
    \STATE {\bfseries Function:}  ServerSide
        \STATE \hspace{0.4cm} $\{u\} \leftarrow$ client\_assigment() \hspace{1.5cm}   \textit{// use  Alg.~\ref{alg:client_assigment} to get client assignment into groups}
        \STATE \hspace{0.4cm}  receive $(w^0_{p2_a};w^0_{p2_b})$ from aggregator
        
        \STATE \hspace{0.4cm} {\bfseries for} {$r=1$ {\bfseries to} $R$} 
        \STATE \hspace{0.7cm} $w^r_{p2_b}[u_g] \leftarrow \{ w^r_{p2_b}  \textit{, } \forall g \in [1,..,G]\})$ \hspace{1.5cm}   \textit{// replicate part-2b for every group}
        \STATE \hspace{0.8cm}{\bfseries repeat} 
        \STATE \hspace{1.2cm} $a^{p1}_{rbc}, y_{rbc} \leftarrow proc\_order() $ \hspace{0cm}   \textit{// use the proc. order policy \& get activations from client $c$}
        \STATE \hspace{1.2cm} $a^{p2_a}_{rbc} \leftarrow w^r_{p2_a}(a^{p1}_{rbc})$ \hspace{1.2cm}   \textit{// fwd-prop on the shared part-2a} 

        \STATE \hspace{1.2cm} $\hat{y}_{rbc} \leftarrow w^r_{p2_b}[u_{c}](a^{p2_a}_{rbc})$ \hspace{1.2cm}   \textit{// find part-2b that client $c$ belongs to \& fwd-prop the activations} 

        \STATE \hspace{1.2cm}  $loss \leftarrow \nabla(\hat{y}_{rbc} - y_{rbc})$   \hspace{0.5cm} \textit{// compute the loss}
        \STATE \hspace{1.2cm} compute $gr^{p1}_{rbc}$ \hspace{0cm} \textit{// compute the gradients of the first cut layer with backward propagation}
        \STATE \hspace{1.2cm} send $(gr^{p1}_{rbc})$ to client $c$  
          \STATE \hspace{1.2cm} update $(w^r_{p2_a};w^r_{p2_b})$ using $loss$ \hspace{0.5cm} \textit{// update the weights of part-2a using the computed gradients from the loss}
        \STATE  \hspace{0.8cm} {\bfseries until} end of batch updates
        \STATE  \hspace{0.8cm} $w^{r+1}_{p2_b} \leftarrow$ aggr($\{w^{r}_{p2_b}[u_g] \textit{, } \forall g \in [1,..,G]\})$ \hspace{0.2cm}   \textit{//aggregate all weights updates from all part-2b heads}
        \STATE  \hspace{0.4cm} {\bfseries end for}
    \STATE {\bfseries EndFunction} 
     \vspace{0.2cm}
   \STATE {\bfseries Function:}  ClientSide (c)
        \STATE $send\_stats(\rho_c)$ \hspace{0.4cm} \textit{// send label statistics}
        \STATE \hspace{0.4cm} {\bfseries for} {$r=1$ {\bfseries to} $R$} 
            \STATE \hspace{0.8cm}  receive $w^r_{p1}$ from aggregator
            \STATE \hspace{0.8cm} {\bfseries for} {$b \in \mathcal{B}$ } 
            \STATE \hspace{1.2cm} $a^{p1}_{rb} \leftarrow w^r_{p1}(b)$
            \STATE \hspace{1.2cm} send $a^{p1}_{rb} , y_{rb}$ to server \hspace{1.5cm}   \textit{// send activations of first cut layer ($a^{p1}_{rb}$) and the corresponding labels ($y_{rb}$)}
            \STATE \hspace{1.2cm} get $gr^{p1}_{rb}$ from server \hspace{1.5cm}   \textit{// get gradients from server}
            \STATE \hspace{1.2cm} $w^r_{p1} \leftarrow w^r_{p1} + \beta gr^{p1}_{rb}$
            \STATE  \hspace{0.8cm} {\bfseries end for}
            \STATE \hspace{0.8cm} send $w^r_{p1}$ to aggregator
        \STATE  \hspace{0.4cm} {\bfseries end for}
    \STATE {\bfseries EndFunction}

\end{algorithmic}
\end{algorithm*}

 \clearpage

The assignment of clients into groups considers two factors, 
\textit{(factor-i)}~the data distribution of the clients and, specifically, the representation of the different labels in the clients'  data, and \textit{(factor-ii)}~repartition of clients among the groups, i.e., ensuring that all heads receive a uniform number of updates (or, in other words, the groups have a balanced number of assigned clients). 
Regarding \textit{(factor-i)}, the goal is to group clients with similar data distributions. In this way, we avoid generating large gradients in the higher layers during iterations through clients. Recall that large gradients impose larger weight changes, which lead to overwriting previous updates, i.e., forgetting the previous knowledge. As shown in~\citep{ramasesh2020anatomy}~(and verified in our observations), the higher layers are responsible for the large gradients that are back-propagated across the whole model. On the contrary, maintaining homogeneity within a group ensures a smooth transition between clients, minimizing significant weight changes and effectively addressing the primary challenge of CF. 

  %\clearpage

\begin{algorithm*}[h]
   \caption{Client assignment to groups~(grouping algorithm)}
   \label{alg:client_assigment}
\begin{algorithmic}
   \STATE {\bfseries Input:} num. of groups \textbf{$G$}, set of clients 
   \textbf{$\mathcal{C}$}
   \STATE {\bfseries Output:} $\{u\}$

   % \STATE
   \vspace{0.1cm}
   \STATE Initialize $\{\mathbf{\rho_c}\} \forall c \in \mathcal{C} \leftarrow$  $get\_stats()$ \hspace{1.5cm} \textit{// get label statistics from clients}
   \STATE Initialize $u_{cg} \leftarrow 0 \text{, } \forall c \in \mathcal{C} \text{, } g \in [1, .., G]$ \hspace{1.5cm} \textit{// start with no clients assigned to any group}
   \STATE Initialize $\mathcal{Q} = \{ c \in \mathcal{C}$ \} \hspace{3.cm} \textit{// define a set with clients who have not been assigned}
   \REPEAT
       \FOR{$g=1$ {\bfseries to} $G$}
       
           \STATE $nextClient \leftarrow max(\{\rho_{cg} \text{, } \forall c \in \mathcal{Q} \}$)
           \STATE $u_{nextClient,g} \leftarrow 1$ \hspace{3cm} \textit{// assign client $nextClient$ to group $g$}
            \STATE $\mathcal{Q} \leftarrow \mathcal{Q} - c$
       \ENDFOR
   \UNTIL{$\mathcal{Q}$ is $empty$}
\end{algorithmic}
\end{algorithm*}

Based on the discussion above, we propose the following approach for grouping the clients into groups and mapping them into different part-2b heads (as shown in Figure~\ref{fig:medusa_workflow}). 
Let $\rho_{cl}$ be the number of samples of label $l$ in client $c$, where $l\in \mathcal{L}$ and $c\in \mathcal{C}$, and 
$\rho_{c} = \{ \rho_{cl} | l \in \mathcal{L} \}, \forall c \in \mathcal{C}$ be  the corresponding vector for each client $c$. 
Furthermore, we define the binary decision variables $\mathbf{u} = \{u_{cg} \mid \forall c \in \mathcal{C}, \forall g \in \mathcal{G} \}$, where $u_{cg} = 1$ when client $c$ is assigned to group $g$, and $u_{cg} = 0$ otherwise.  By design, each client must be assigned to exactly one group, i.e., 
\begin{equation} \label{eq:constraint-alg2}
    \sum_{g \in \mathcal{G}} u_{cg} = 1, \forall c \in \mathcal{C}.
\end{equation}

Finally, considering the \textit{(factor-i)}, and \textit{(factor-ii)} discussed earlier, we adopt the max-min approach (from the theory related  to resource allocation problems) and formulate the following objective:
\begin{equation}
\label{eq:objective_group}
\max \min_{g \in \mathcal{G}}  \frac{\sum_{c \in \mathcal{C}} \rho_{cl} u_{cg}}{\sum_{c \in \mathcal{C}} u_{cg}}, \forall c \in \mathcal{C}
\end{equation}

\noindent subject to the constraints in~\eqref{eq:constraint-alg2}.
We propose a heuristic for the optimization problem above in Algorithm~\ref{alg:client_assigment}. In a nutshell, Algorithm~\ref{alg:client_assigment} decides the assignments of clients to groups in a greedy way. In detail, for each group, the algorithm finds the client with the largest number of samples from the corresponding label, according to \textit{(factor-i)}. The algorithm repeats this process until all clients have been assigned to exactly one group; satisfying the constraint discussed above.
Finally, we note that Algorithm~\ref{alg:client_assigment} iterates through the groups and assigns to them clients that are not assigned yet to any group. By doing so, and not iterating through the clients instead, the repartition of clients into the groups is balanced, considering \textit{(factor-ii)}.

The complexity of Algorithm~\ref{alg:client_assigment} is equal to $O(G \cdot C)$, because, for $G$ iterations, each of them concerning a group $g\in \mathcal{G}$, we find the maximum element of the vector $\{\rho_{cg} | c\in \mathcal{C}\}$, where the complexity of finding the maximum of a vector/array of length $C$ is equal to $O(C)$. It is important to stress that the computing overhead of $O(G \cdot C)$ of Algorithm~\ref{alg:client_assigment} is a one-time cost, since it only needs to run at the beginning of the training. Finally, we note that one could devise an alternative algorithm or even solve exactly the optimization problem in Eq.~\eqref{eq:objective_group} (e.g., with an optimization solver), but at the cost of higher computational complexity.

\section{\revision{Variants of Split (Federated) Learning}}
\label{ap:protocols}

\begin{figure*}[h]
    \centering
\includegraphics[width=0.9\textwidth]{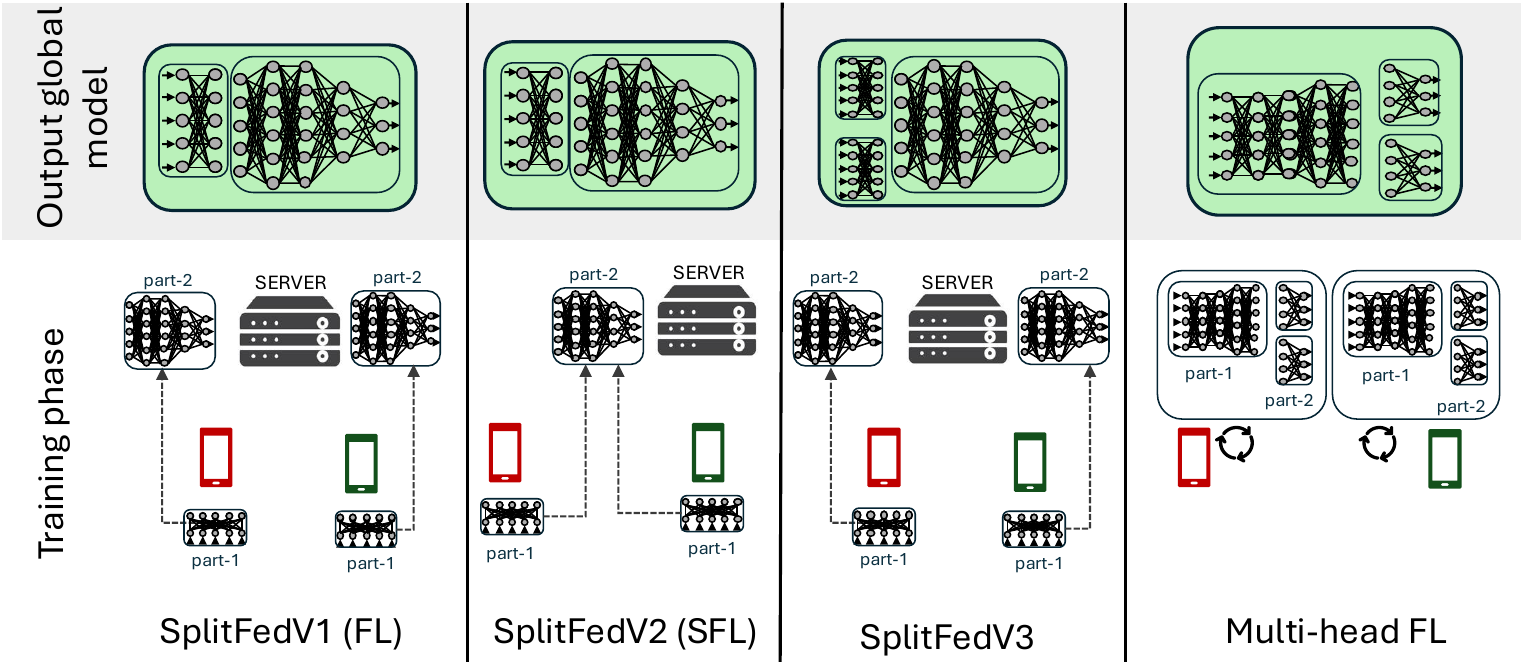}
\caption{\revision{A comparative overview of different distributed learning protocols in terms of model parts used during training and at the end of the training (output model).}}
\label{fig:other_algos}
\end{figure*}

\revision{There is a plethora of proposed definitions for SL, e.g., SplitNN~\citep{vepakomma2018split} or parallel SL~\citep{jeon2020privacy}, 
that mostly differ in the way
the training at the server takes place and the presence of an aggregator. Among them, 
SplitFed~\citep{thapa2022splitfed}~(also called Split Federated Learning\footnote{\revision{We clarify that, for simplicity, the abbreviation ``SFL'' always refers to SplitFedv2 throughout this manuscript, unless otherwise specified.}}) is an emerging method for training deep Neural Network models in a system comprised of multiple IoT and mobile devices. 
We introduce here the most widely used variants of SplitFed and discuss their differences. 
Figure~\ref{fig:other_algos} provides an overview of each variant of SplitFed, as well as the multi-head FL paradigm~\citep{chen2023flexibility}. We note that the training steps~(forward/backward propagation) for the three variants of SplitFed~(i.e., versions 1 to 3) are similar to the ones depicted in Figure~\ref{fig:intro_pic}. Moreover, in all three variants, part-1 is handled by the clients and part-2 is handled by the server.}

\smallskip \noindent \textbf{SplitFedV1} was the first SplitFed  variant proposed in~\citep{thapa2022splitfed}, also often referred as Parallel SL~\citep{tirana2024workflow}. 
As shown in Figure~\ref{fig:other_algos}, there is a different version of the model parts for each client.  
This enables the distinct segments of the model to be trained entirely in parallel, replicating a paradigm similar to the one of FL. Additionally, the model parts are aggregated at the end of each round, and the global model is the result of concatenating the two model parts.
The main drawback of this setup is the computing and memory demands on the server side, i.e., the server preserves the state of part-2 for every participating client. Moreover,  the fact that the model convergence guarantees are the same as in FL, and not as in conventional SL, i.e., SplitNN~\citep{vepakomma2018split}, can be considered another drawback. This is because often, as we have seen in the literature, SL paradigms can outperform FL~(which is also shown in Figure~\ref{fig:scatter_plot} in the case of IID data distribution).

\smallskip \noindent \textbf{SplitFedV2~(SFL)} is the second variant of SplitFed  introduced in~\citep{thapa2022splitfed}. This version aligns more  
with the original (sequential) Split Learning (SL) framework, i.e., SplitNN~\citep{vepakomma2018split}, when compared to SplitFedV1. Unlike SplitFedV1, which operates with fully parallel model updates, SFL incorporates a shared model segment, specifically part-2, among the clients. Consequently, clients perform parallel updates on part-1, while part-2 is updated sequentially.
The advantages of this approach include: \textit{(i)} enhanced model accuracy relative to SplitFedV1~(FL), as shown in~\citep{thapa2022splitfed} and \textit{(ii)} reduced server resource requirements compared to SplitFedV1. However, as discussed in this paper, SFL is susceptible to catastrophic forgetting (CF), a phenomenon that is further pronounced under heterogeneous data distributions among the clients.

\smallskip \noindent \textbf{SplitFedV3} was introduced ~\citep{gawali2021comparison} and \citep{madaan2022vulnerability}, and was specifically designed to address CF~\citep{madaan2022vulnerability} in Split Federated Learning~(SFL). The training process closely mirrors that of SplitFedV1, where each client maintains its version of part-1 and part-2. However, at the end of each round, only part-2 is aggregated. Consequently, the final model features multiple input layers, corresponding to the layers that comprise part-1. \revision{Nevertheless, the process for selecting a part-1 during inference is not explicitly defined. To this end, we follow the Multi-head~\citep{chen2023flexibility} approach and choose the part with the most confident output (see below).}

\smallskip \noindent \textbf{Multi-head FL} was proposed in~\citep{chen2023flexibility} and incorporated the multi-head approach into FL. Specifically, it considers a continual learning~(CL) setup with incremental learning. Hence, a multi-head neural network architecture with task-specific layers is proposed to capture and preserve unique features for each task, enhancing prediction accuracy and model robustness. During the training phase, each client trains the whole model and the multi-heads locally. Then, at the end of each round, all layers are aggregated while preserving the multi-heads~(i.e., the head $x$ of client $c$ is aggregated with the corresponding head from the other clients). As a result, the final model is comprised of two parts, the main body, shared for all tasks, and the different head layers. In order to make predictions during inference~(when training has been completed), the following formula is used to make the selection between the head layers, 
\begin{equation}
y_{pred} = \argmax_{b \in \mathcal{B}} f_b(x;w_b),
\end{equation}
where $f_b(x;w_b)$, represents the model trained with task $b$ with parameters $w_b$. This function selects the label corresponding to the highest prediction value among all the task-specific models.

\section{Extended Related Work}
\label{ap:ext_related_work}

In this section, we expand the Related Work section of the main paper. In Section~\ref{sec:related_work}, we discussed works concerning Federated Learning and Split Learning settings, which are more related to this manuscript's focus, while in this section, we present works on  Continual Learning~(CL) that endeavor to tackle Catastrophic Forgetting~(CF). Then, we provide a more thorough discussion on regularization-based mitigation methods.

\subsection{Catastrophic Forgetting in Continual Learning}

The goal of CL is to learn through a dynamic data distribution, and forgetting is measured by the metric of backward transfer, which estimates the impact of learning a task on the knowledge gained through the old tasks~\citep{wang2024comprehensive}. 
Mitigation techniques usually fall under  5 categories: regularization-based 
(discussed in Appendix~\ref{ap:reg_CL}), 
replay-based, 
optimization-based, representation-based, and architecture-based.  
Works that provide insights on the phenomenon of forgetting through systematic analysis and propose mitigation techniques include \citep{ramasesh2020anatomy,ramasesh2021effect, kemker2018measuring}. 
An aspect that is less studied in the literature, but relevant to our topic, is the impact of repetitions or cyclic order of the input data. In fact, repetitions can alleviate forgetting~\citep{stojanov2019incremental, cossu2022class}, and mitigation techniques based on multiple (sub-)~networks per task/experience have proven effective against forgetting~\citep{hemati2025continual}.
Theoretical results on cyclic learning for linear tasks are provided in~\citep{evron2023continual, swartworth2023nearly}. Another direction is to study CF with respect to the order of the input data in the data stream. Theoretical works focusing on linear models include \citep{lin2023theory}. 
Finally, although not the focus of this manuscript, Curriculum Learning~\citep{bengio2009curriculum} studies the order of training samples with respect to their semantic relationship.
\subsection{Regularization in Continual Learning}
\label{ap:reg_CL}

In general, the concept of regularization involves adding a penalty term to the loss function, which restricts the weights from increasing (limiting the large changes in the weights). A common regularization technique in Machine Learning is L1 and L2 regularization, which is used to avoid model overfitting. Specifically, in the CL literature, the idea of regularization is extended as a way to balance old and new tasks by storing important model weights before altering them. One such example is 
EWC~\citep{kirkpatrick2017overcoming}, which calculates the importance of weights using the Fisher information matrix~(FIM) and, according to that, adds a quadratic penalty in the loss function.

However, such approaches cannot be directly applied in SFL, since they are inherently linked to the notions of ``old" and ``new" data~(in the setting of CL the data stream is temporal). In fact, such approaches require the model to be first fully trained on one of the tasks and then compute the FIM before proceeding to the next task. But in SFL, the model is updated by processing all tasks in every round, as explained in Section~\ref{sec:intro}. That is because the server receives successive updates from all clients~(in this work, we do not consider client selection policies). 
Finally, in contrast to CL, in SFL, no single entity in the system has access to or control over the entire model during training. We remind that, in SFL,  the clients (and the aggregator) manage part-1 of the model, while the server manages part-2.

\section{Further Details on the Experimental Setup}\label{ap:implem_details}

\subsection{\textbf{Implementation Details}} 
\label{ap:implem_details_1}
\noindent \textbf{Training Setup.} We study the performance of the models after training for $100$ training rounds. All models are trained from scratch -- no pre-trained models are used. We use the stochastic gradient descent~(SGD) optimizer, with a learning rate equal to $0.05$ and a decay rate equal to $0.993$~(with min lr=$0.005$), while the batch is equal to $64$ data samples. The code is implemented using the PyTorch library\footnote{\url{https://pytorch.org/}} and the datasets from the ``Hugging Face''\footnote{\url{https://huggingface.co/}} website. For the data partitioning, we utilize the well-known FLOWER~\footnote{\url{https://flower.ai/docs/framework/index.html}} framework. 
Even though we use open-source models, we need to re-implement them~(i.e., code manually the model's classes) to incorporate the SL operation. In the experiments, we use $L$ clients, where $L$ is the total number of labels (classes) that the dataset has. However, note that the number of clients is increased to simulate more challenging scenarios. In detail, this is annotated with the multiplier $\phi$, hence, we employ experiments with $L\cdot \phi$ clients. 
More details are given in %\textbf{our source code, which is provided in the Supplementary Material of the submission and will become available upon publication.}
our source code.\footnote{\url{https://github.com/jtirana98/Hydra-CF-in-SFL}}
Finally, to run all experiments, we use a private cluster with multiple machines, which allows us to run multiple instances of the same experiment, but with different seeds. In detail, the hardware characteristics of these machines are: $2\times$ Xeon $6140$  ($2.3$Ghz, $18$ Cores) \& $2$ Nvidia Tesla V100 ($32$GB) and machines with NVIDIA A$100$-SXM4-$40$GB. 

\noindent \textbf{Repeated Experiments and Statistical Significance.} All experiments are repeated at least $10$ times\footnote{The only exception is when we study a single experiment, where we explicitly specify that it is a snapshot of a single experiment, as in Appendix~\ref{ap:conf_matrix}, for example.}~(changing seed). This is a common practice
to avoid making assessments based on one experiment, see, e.g.,~\citep{li2022federated}. We note that variations in experimental results may occur due to weight initialization, stochastic training parameters, etc. 
As a result, throughout the paper, we report the median of the repeated experiments so that our results are robust against outliers~\citep{leys2013detecting}. Since historically, outliers were detected through mean-based metrics~\citep{leys2013detecting}, we also include the standard deviation. Specifically, at the end of each round, we compute the accuracy of the global model~(consisting of the aggregated part-1 and server's part-2), the BW, and the PG, based on the per-label accuracies. Then, for each quantity, the median and standard deviation from all repeated experiments are computed.

\subsection{Cut Layer Selection}
\label{ap:cut_layer}

The cut layer  is one of the key parameters in SL protocols 
\revision{since, as we show in our analysis, it affects the performance of the model when the training contains sequential updates between the clients~(as in, e.g., SplitNN~\citep{vepakomma2018split} and SFL) and the impact of catastrophic forgetting. Moreover, the cut layer selection has an impact on the training time which, however, is not the focus of this work.}

Considering the corresponding literature on how the cut layer selection is made, we notice that there are different approaches. Firstly, 
the authors in~\citep{thapa2022splitfed} selects arbitrarily relatively shallow cut layers, i.e., the second layer of LeNet~(after the 2D MaxPool layer), the second layer of AlexNet~\citep{krizhevsky2012imagenet} (after the 2D MaxPool layer), the fourth layer of VGG16~\citep{simonyan2014very} (after the 2D MaxPool layer), and the third layer (after the 2D BatchNormalization layer) of ResNet18~\citep{he2016deep}. \revision{A similar approach is observed in~\citep{liu2022wireless}.} 
For consistency throughout this manuscript, we consider the bottleneck search selection, which is common and widely used. This selection method was first introduced in~\citep{kang2017neurosurgeon} and has been applied in plenty of works of split computing, in general.

Finally, we note that the cut layer does not play a crucial role in SplitFedV1 in terms of performance. As we discussed in Appendix~\ref{ap:protocols}, in SplitFedV1, all clients have a different version of the model parts that are trained in parallel (part-1 at the client and part-2 at the server). As a result, the performance of SplitFedV1 is the same as FL independently of the cut layer~\citep{thapa2022splitfed}.

\section{Simulating Data Heterogeneity Patterns}
\label{ap:sim_data_hetero}

Previous studies~\citep{kairouz2021advances, zhu2021federated, solans2024non} provide a taxonomy of the state-of-the-art strategies used to create non-IID data distributions for distributed learning research.
One of the most prominent approaches among the main categories in the provided taxonomies corresponds to Label Distribution Skew, which implies that the label  distribution $P_i(y)$ varies among clients while the conditional label-feature distribution $P_i(x | y)$ remains unchanged.
Label distribution skewness seems to be the most challenging one~\citep{li2022federated,liao2023mergesfl}, 
especially when the clients have a restricted view of the training task. 
As discussed in previous sections of this manuscript, we conduct our experiments using Label Distribution Skew scenarios. 
In the upcoming subsections, we provide further details for each of the Label skew partitioning approaches that have been used.
 
\subsection{Dominant Label Ratio}
\label{ap:sim_data_DL}

\begin{figure}[!h]
    \centering
\includegraphics[width=1.\columnwidth]{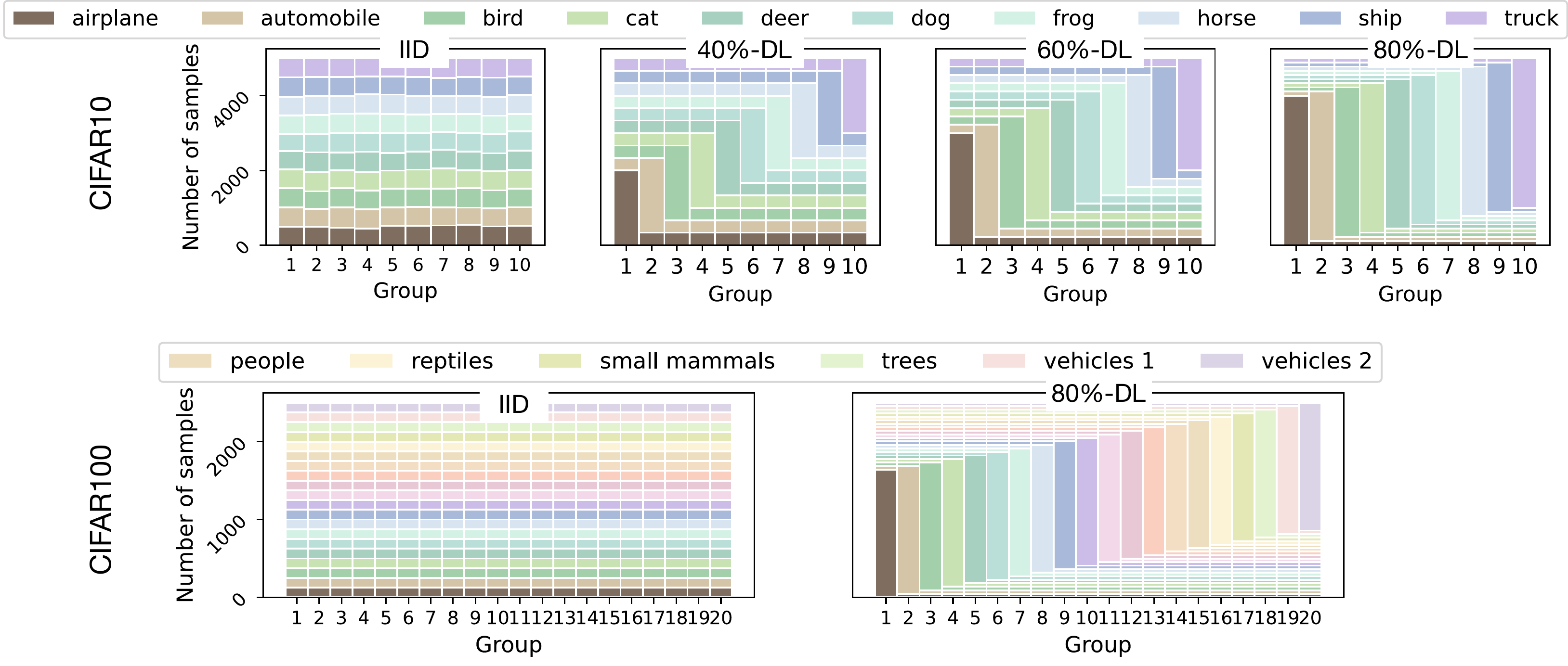}
\caption{Visualization of clients' data distributions obtained using the \textit{Dominant label ratio} approach for CIFAR-10 (top) and CIFAR-100 (bottom). 
For visualization purposes, we restrict the legend of CIFAR-100 to a subset of label names.
}
\label{fig:non_iid_sim}
\end{figure}

This approach ensures that clients have examples from all labels; however, they possess a disproportionately higher number of samples for a particular label, referred to as the dominant label. It uses the parameter~$p \in [0,100]$ to control the percentage~(i.e., $p\%$) of a dominant label at each client, while the remaining samples~(i.e., $(100-p)\%$) are distributed evenly among the other labels. Clearly, as $p$ increases, the data partitioning gets more heterogeneous among the clients. Figure~\ref{fig:non_iid_sim} provides an overview of the various clients' data distributions created with this method with various $p$ values.

\subsection{Dirichlet Partitioning Method}
\label{ap:sim_data_Dirichlet}
Dirichlet partitioning is probably the most commonly used method to federate data in distributed computing research~\citep{solans2024non}. It leverages the Dirichlet Distribution~\citep{minka2000dirichlet} to allocate samples among clients, effectively simulating non-IID scenarios. The concentration parameter, $\alpha$, determines the level of data heterogeneity: lower values result in more imbalanced distributions, while higher values produce distributions closer to IID conditions. 
This approach enables the creation of diverse non-IID datasets, facilitating the evaluation of FL algorithms under varying levels of data imbalance.

\subsection{Custom Sharding Method}
 \label{ap:sim_data_shard}
The sharding technique \citep{zhang2021federated} generates non-IID datasets by first sorting data based on class labels and dividing it into smaller, equally sized subsets known as \textit{shards}. These shards are then distributed to clients, with each client receiving a fixed number. As each shard contains data from only a few classes, clients end up with datasets representing only a limited portion of the overall class distribution.
 This approach creates a highly skewed, non-IID distribution, where clients predominantly have samples from a small number of classes, without any overlap in class representation across clients from different shards.

Starting from this methodology, we develop a custom approach that uses the notion of shards combined with class overlapping among them. 
To do so, the original dataset is divided into $L$ groups, each of which contains a specific subset of $n \geq 2$ labels as dominant labels per group.
The distribution of samples within each group is governed by the non-IID parameter, denoted as $p\%$. Specifically, $\frac{p}{n}\%$ describes the samples belonging to a given label $l \in L$ that are assigned to the client belonging to the group where $l \in L$ is the dominant label, while the remaining $(100 - p)\%$ of the samples are evenly distributed across the other groups.

\section{Types of Processing Orders}
\label{ap:other_orders}

As we discussed in Section~\ref{sec:order} the existing literature of Split Federated Learning (SFL), does not consider a specific processing order on the server. Specifically, they assume a random order which serves clients' request as they arrive at the server~(i.e., first-come-first-served). Here we present two additional variances, which are more structured.

\begin{figure*}[th]
    \centering
\includegraphics[clip, width=0.8\columnwidth, trim={0cm 1.3cm 0cm 2cm}]{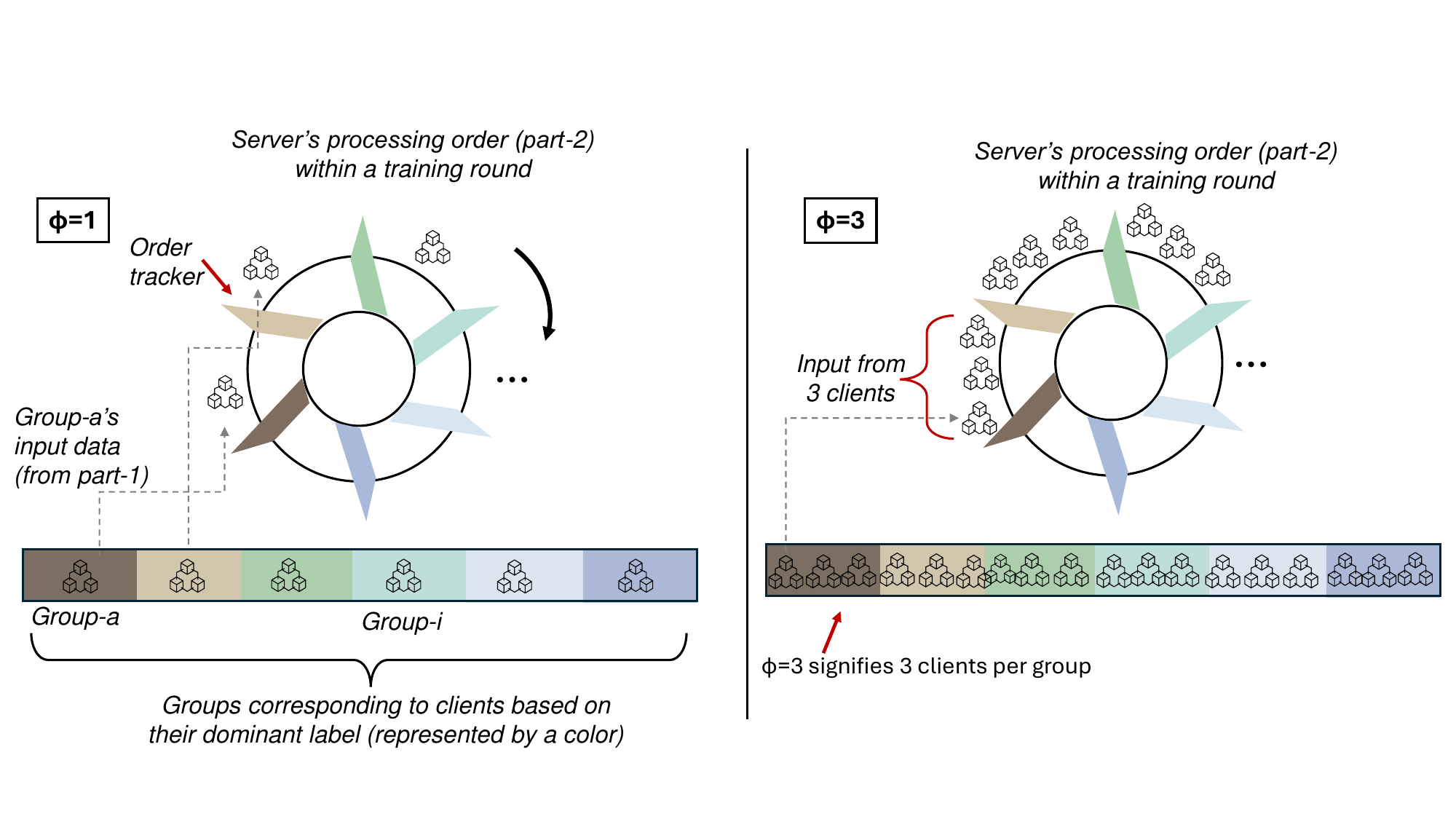}
\caption{An abstract illustration of \textit{cyclic order} for 2 different values of $\phi$ and $L=6$ labels (each represented by a different color).}
\label{fig:cyclic-order}
\end{figure*}

\subsection{Cyclic Order}
\label{ap:cyclic_order}

Figure~\ref{fig:cyclic-order} presents an illustration of the cyclic order. In detail, before the actual training starts the clients sharing the same Dominant Label~(DL) are arranged according to a predefined order. Then the server during training iterates cyclically through the $L$ groups. The server selects all clients of one group before proceeding to the next one.

\subsection{Cyclic-and-reverse Order}
\label{ap:permutation} 

For this policy, there is a pre-defined order of the group of clients, like in cyclic order, but in every $\phi \cdot L$ steps, this order is reversed. In other words, one can think of it as if the ``wheel'' in Figure~\ref{fig:cyclic-order} is successively rotated right and left, repetitively. The intuition behind this order derives from the observation that the dominant label of the last-processed client in cyclic order achieves better accuracy than the dominant labels of the clients processed earlier in the sequence, as we discussed in Section~\ref{sec:order}.

\section{Extended Systematic Analysis of Catastrophic Forgetting in SFL}
\label{ap:ext_syst_analasis}

In this part, we provide supplementary results of the empirical analysis conducted in Section~\ref{sec:intraCF}.

\subsection{Performance under Other Data Partitions}
\label{ap:syst_other_data_pat}

\begin{table*}[!th]  \caption{Global accuracy~(reported median of the last five rounds across all runs) and PG of SFL for other data partitions, supplementary to Table~\ref{tab:unfairscore_cyclic}. The results presented here are for cyclic order with $\phi=10$. The parentheses present the global accuracy and PG, respectively, for IID.}
\label{tab:other_data_pattern_systematic}
\centering 
\centering 
\begin{tabular}{p{1.7cm}p{1.2cm}|p{1.5cm}p{1.2cm}p{1.2cm}|p{1.5cm}p{1.4cm}p{1.2cm}}
 &  & \multicolumn{3}{c|}{ \textbf{\small Global Accuracy $\uparrow$}} & \multicolumn{3}{c}{ \textbf{\small{Performance Gap $\downarrow$}} }\\ 
 
 \textbf{Model} & \textbf{IID} & Sharding & \small Dir.$_{\alpha=0.3}$ &  \small 
 Dir.$_{\alpha=0.1}$ & \small Sharding & \small Dir.$_{\alpha=0.3}$ &  \small Dir.$_{\alpha=0.1}$ \\ \hline

   & & \multicolumn{6}{c}{{\textbf{CIFAR-10 $(\mathbf{K=10})$}}} \\ 
\multirow[c]{1}{*}{MobileNet} & (80, 10)  & $\mathbf{50}$ \small{$\mathbf{\pm 0.4}$} &  \textbf{50} \small{$\mathbf{\pm 0.5}$} & $40$ \small{$\pm 0.6$} &  \textbf{36} \small{$\mathbf{\pm 1}$} & $43.7$ \small{$\pm 0.5$} & $45.5$ \small{$\pm 1$}\\
\hline
\multirow[c]{1}{*}{ReNet101} & (68, 15)  & \textbf{42} \small{$\mathbf{\pm 1.2}$} & $33$ \small{$\pm 0.6$} & $27$ \small{$\pm 1$} & \textbf{35} \small{$\mathbf{\pm 3}$} & $53$ \small{$\pm 2$} & $55$ \small{$\pm 3$}\\ \hline

 & & \multicolumn{6}{c}{{\textbf{TinyImageNet} $(\mathbf{K=200})$}} \\ 
\multirow[c]{1}{*}{MobileNet} & (50, 40)  & $27$ \small{$\pm 0.3$} & \textbf{45} \small{$\mathbf{\pm 0.3}$} &  $31$ \small{$\pm 1$} & $62$ \small{$\pm 1.6$} & \textbf{46} \small{$\mathbf{\pm 1}$} & $63$ \small{$\pm 0.7$}\\
\hline
\end{tabular}
\end{table*}

Table~\ref{tab:other_data_pattern_systematic} presents the performance (PG and global accuracy) of SFL using cyclic order with $\phi=10$ and additional data partitions to the one shown in Section~\ref{sec:intraCF} in Table~\ref{tab:unfairscore_cyclic}. In detail, here we consider the sharding and Dirichlet data partitioning, note that for the latter, as the $\alpha$ parameter decreases, the more non-IID the data distribution becomes. The table shows that for CIFAR-10, both models' performance deteriorates. Specifically, CIFAR-10 and Dirichlet with $\alpha=0.1$ challenges the most, both models, which is also true for TinyImageNet, e.g., has a $57\%$ increase in the PG compared to the IID case.

\subsection{Snapshot of Experiments and Confusion Matrix}
\label{ap:conf_matrix}

Firstly, we study a snapshot of a single experiment for the two types of processing order, 

\begin{figure*}[h]
    \centering
\includegraphics[width=0.95\columnwidth]{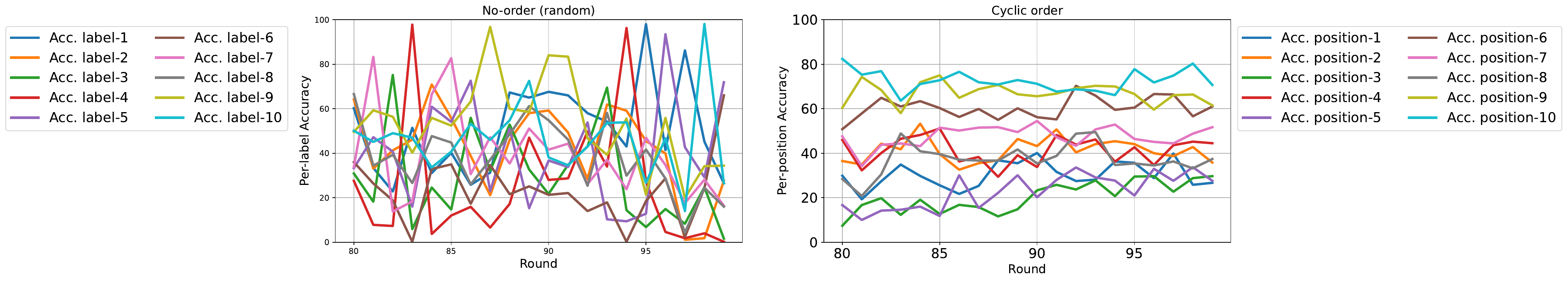}
\caption{Snapshot of per-label accuracy for a single experiment, training MobileNet with CIFAR-10 and $80\%$-DL. The left part shows an instance from random processing order, while the right part shows the result of cyclic order with $\phi=10$.} 
\label{fig:snapshot}
\end{figure*}

\noindent  $\bullet$ \textbf{Random order.} The left part of Figure~\ref{fig:snapshot} shows a snapshot of a single experiment for the no-order policy (random), in which the server selects the clients in a non-predefined manner. 
We notice that the training is unstable even after several training rounds, since all per-label accuracies demonstrate intense fluctuations. In detail, the plot shows that for every per-label accuracy, there are extreme peaks and falls from round to round, i.e., at one round the per-label accuracy is $100\%$, and in the next one drops dramatically. \\
$\bullet$ \textbf{Cyclic order.} Figure~\ref{fig:intro_nonIID} in the main text presents a snapshot of one experiment with cyclic order and $\phi=1$. At first glance, we see that the training seems more stable compared to the snapshot of random order, e.g., the per-label accuracies for all labels throughout training increase in a smoother manner and with fewer fluctuations. 
However, as discussed in Section~\ref{sec:intro}, a disparity in the performance can be seen as the labels served last get better accuracy. Specifically, the snapshot of $\phi=10$ on the right of Figure~\ref{fig:snapshot} shows how this phenomenon is amplified even more as $\phi$ increases.

These findings show that for non-IID, the performance drops significantly for both approaches. In detail, the results show that training gets more unstable, i.e., the per-label accuracy for all labels does not converge smoothly at one point when we use the random processing order. On the other hand, in cyclic order, even though the training seems smoother, a disparity between the per-label's accuracies is witnessed~(as also shown in the preliminary results in Figure~\ref{fig:intro_nonIID}).

\begin{figure*}[h]
    \centering
\includegraphics[width=1\columnwidth]{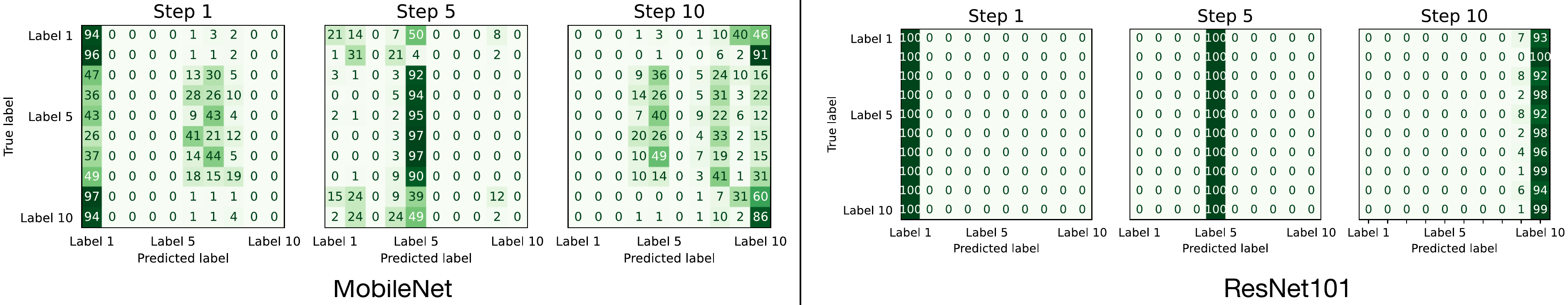}
\caption{Confusion matrix for a snapshot of sequential steps inside one training round. The step corresponds to the group of clients with the same dominant label, in the cyclic order. The plot shows the result for MobileNet and ResNet101 with $\phi=10$ and CIFAR-10.}
\label{fig:conf_matrix_app}
\end{figure*}

\begin{figure*}[h]
    \centering
\includegraphics[width=0.95\columnwidth]{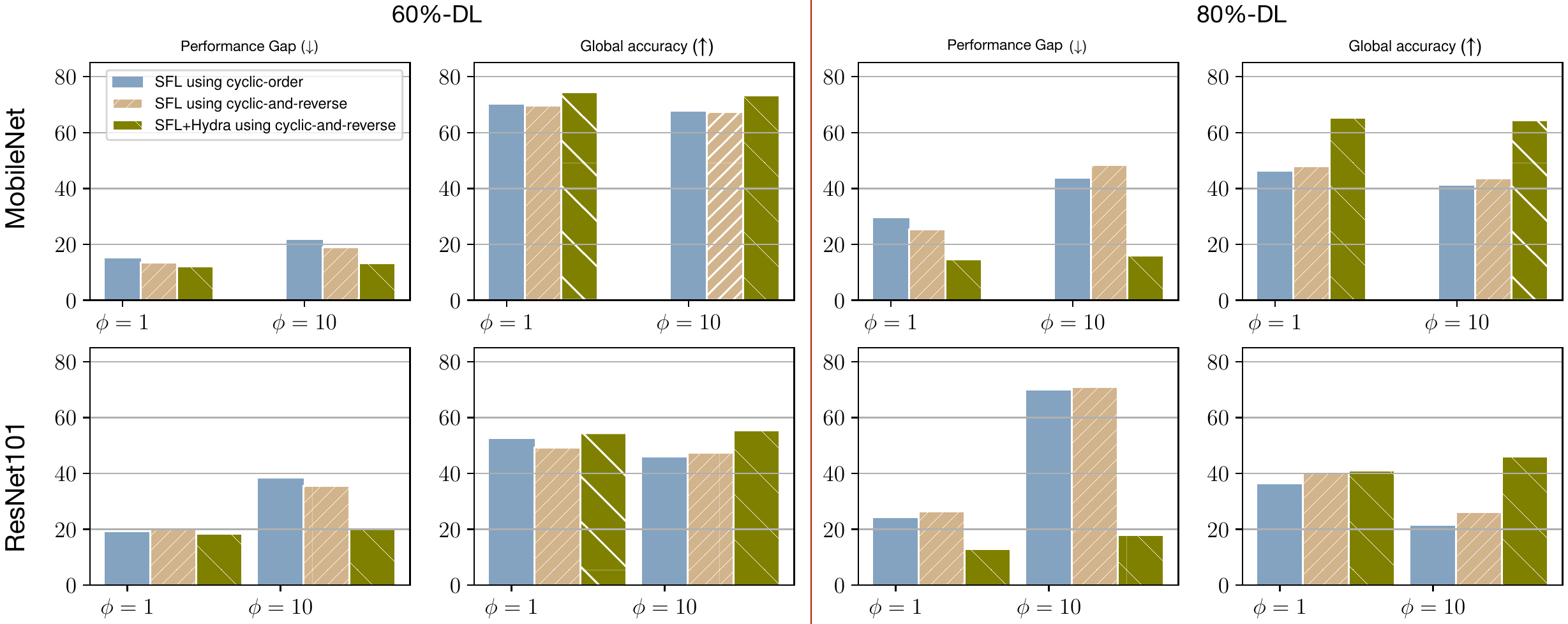}
\caption{Analyzing the \textit{cyclic-and-reverse} order. The plot shows the PG and global accuracy (median of the last five rounds across all runs) for the SFL and Hydra using the cyclic-and-reverse order using CIFAR-10. 
}
\label{fig:permutation}
\end{figure*}

Diving more into the cyclic order,  Figure~\ref{fig:conf_matrix_app} shows the \textbf{confusion matrix} for the labels during an early training round for the MobileNet and ResNet101. 
The step represents the group of clients, with the same dominant label, that most recently updated the model. In detail, in step $x$, $\phi$ clients, whose dominant label is $l_x$, have been updating the model. The figure highlights that the model tends to overfit to the dominant label at each step, \textit{capturing the similarities between intraCF and CF in CL}

\subsection{Studying Other Orders}
\label{ap:test_reverse}

Considering the regular SFL in Figure~\ref{fig:permutation} we compare the performance of SFL with cyclic-and-reverse order and cyclic order. The cyclic-and-reverse order does not always outperform the cyclic order. Firstly, considering the MobileNet for $60\%$-DL, even though the cyclic-and-reverse order manages to reduce the PG, up to $12\%$, there is no improvement in the accuracy. But, for $80\%$-DL and $\phi=10$, the cyclic-and-reverse and cyclic have similar results. Also, a dependency between $\phi$ and the PG is manifested, i.e., the PG increases as the $\phi$ increases.

\subsection{Studying the Per-Position Accuracy}
\label{ap:per_pos_all}

\noindent \textbf{Per-position Accuracy.} %The position refers to the order in which client $c_l$ with dominant label $l$ is processed at the server.  In practice, position-$k$ accuracy, where $k\in \{1 \ldotp \ldotp L\}$, is the accuracy of the labels that are dominant in clients that are processed at position $k$ (in the processing cycle of length $L$). 
The \emph{position} denotes the point in the processing order at which client $c_l$, whose dominant label is $l$, is served by the server. In practice, the \emph{position-$k$} accuracy (for $k\in \{1 \ldotp \ldotp L\}$,) corresponds to the accuracy of the labels that are dominant among the clients processed at position~$k$ within each processing cycle of length~$L$. 
To ensure that the results are \textit{independent} of the type of label, the order of the labels is randomly allocated in every experiment, and then the median value from all repeated experiments is computed to give the per-position accuracy. 
This definition is tailored to the cyclic order and DL partitioning, but can be extended to other types, too. 

In the following, we present supplemental results to the ones presented in Section~\ref{sec:intraCF}. Note that we incorporate three different models, and for each of them, we use three different cut-layer levels~(i.e., shallow-cut, mid-cut, deep-cut). In detail, the definition of these cuts, respectively, for each level, is: (i)~MobileNet has $26$ layers in total and uses cuts $4, 15$, and $23$, (ii) ResNet101 has $37$ layers in total and uses cuts $2, 10$, and $33$, (iii) ResNet18 has $12$ layers in total and uses cuts $2, 6$, and $10$. Note that for the ResNet models we consider that each residual block is a single layer, like in~\citep{tirana2024workflow}. Moreover, recall that the cut-layer selection follows the bottleneck policy as discussed in Appendix~\ref{ap:cut_layer}.

\begin{figure*}[h]
\begin{center}
  \includegraphics[width=1\columnwidth]{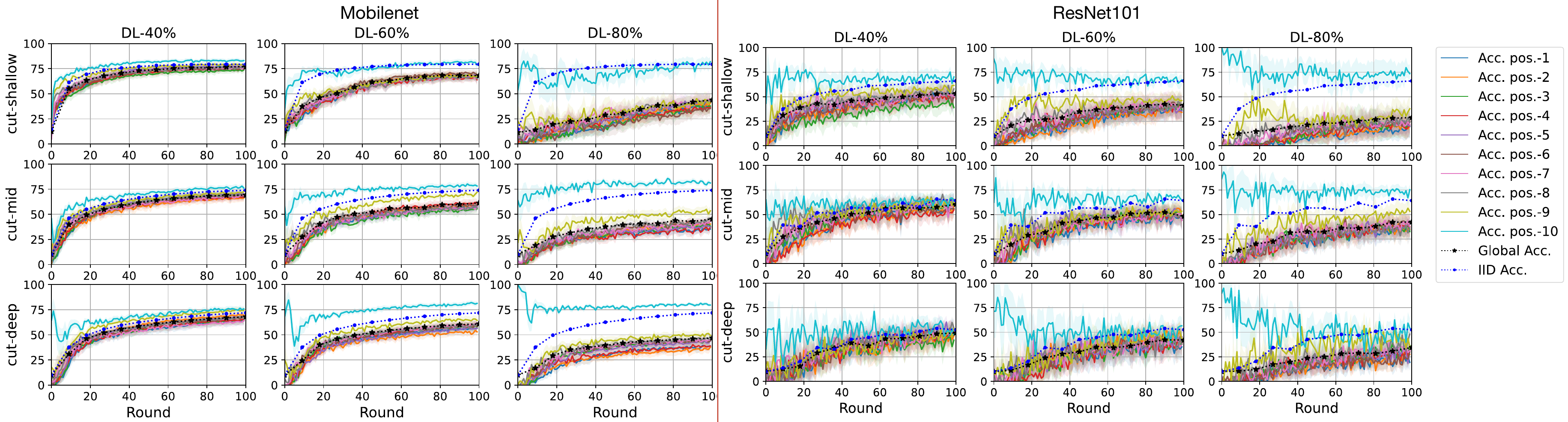}
  
\end{center}
\caption{\revision{The per-position and global accuracy trained on MobileNet (left side) and ResNet101 (right side) using CIFAR-10 for different cut layers and DL ratios (and confidence interval of $95\%$).}
}
\label{fig:cifar10_perposition_app}
\end{figure*}

$\bullet$ \textbf{Figure~\ref{fig:cifar10_perposition_app}:} This plot presents the per-position accuracy for CIFAR-10 trained on models MobileNet and ResNet101. We notice that in both models, as the DL ratio increases, the performance of the model deteriorates in terms of global accuracy, and the distance between the per-position accuracy of position $10$ (i.e., the last position in the cyclic order) and the rest of the labels becomes larger. 
In the case of ResNet101 (right) under $60\%$-DL and $80\%$-DL partitioning, as we go from shallow to deep cut, i.e., from top to bottom: (i) the disparity in per-position accuracies fades, and (ii) the gap between the global accuracy of the non-IID scenario and the one of IID decreases. This is true in most cases, with an exception observed for MobileNet under $80\%$-DL.

\begin{figure*}[!ht]
\begin{center}

  \includegraphics[width=1\columnwidth]{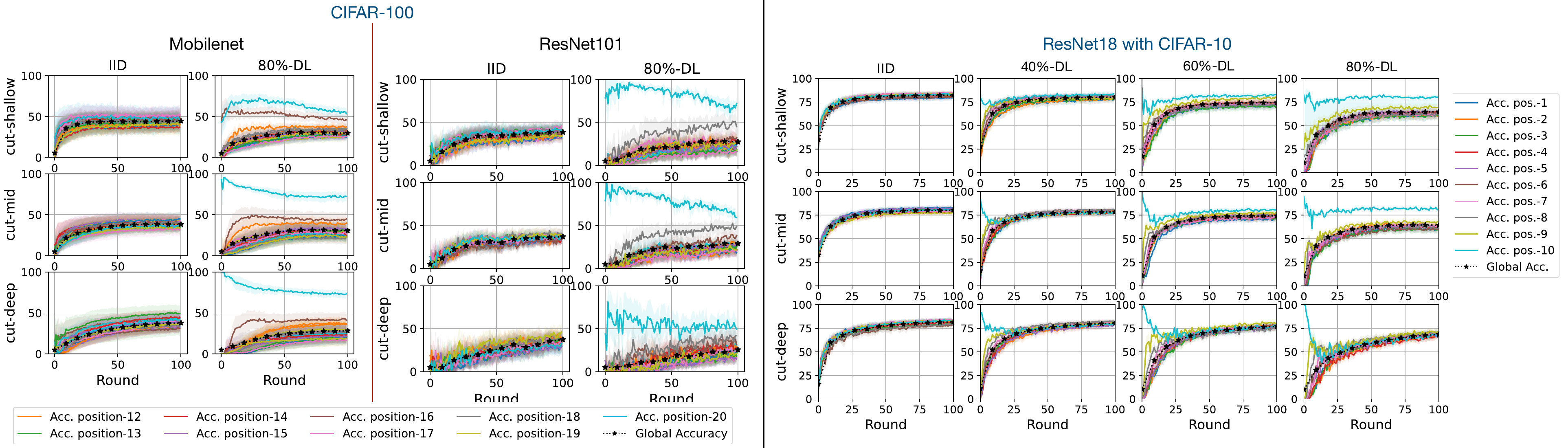}
\end{center}
\caption{The per-position and global accuracy trained on \revision{CIFAR-100 with MobileNet and ResNet101 (left side)}, and CIFAR-10 using ResNet18  (right side). In all cases,  different cut layers and DL ratios are used. For a better visualization, only the results for the last 10 positions are shown for the CIFAR-100. The shaded area around each line represents the confidence interval of $95\%$.
}
\label{fig:CIFAR-100_perposition}
\end{figure*}

\textbf{$\bullet$ Right part of Figure~\ref{fig:CIFAR-100_perposition}:} The plot presents the per-position accuracy of ResNet18 and CIFAR-10. Note that ResNet18 is a less complex version of ResNet, compared to ResNet101. 
Here, intraCF, compared to ResNet101, is less intense for the shallow and middle cut, whereas for the deep cut, the distance is shrunken, like in ResNet101. This is in line with the findings of Table~\ref{tab:unfairscore_cyclic}, which highlight that \textit{larger and more complex models are more prone to catastrophic forgetting than simpler ones}.

\textbf{$\bullet$ Left part of Figure~\ref{fig:CIFAR-100_perposition}:} This plot presents the results for the CIFAR-100 dataset using superclasses, i.e., $L=20$. The plot shows that the accuracy of position $20$ (i.e., last position) is always significantly larger than the others. Moreover, as the DL ratio increases (i.e., from IID to $80\%$-DL), the distance between position $20$ and the other layers is amplified. However, the effect of CF for CIFAR-100 is not as intense as when using CIFAR-10. This is also framed in Table~\ref{tab:unfairscore_cyclic}, in which we observe that \textit{as $L$ decreases, the impact of CF in SFL is greater}, e.g., when comparing the PG of IID and $\phi=1$ in Table~\ref{tab:unfairscore_cyclic}, we see that for CIFAR-10 the increase is up to $\times3$ times larger, and $\times 6.5$ for SVHN, whereas for CIFAR-100, $L=20$ and $100$, the increase is up to $\times 1.4$ and $\times 1.1$, respectively.

\begin{figure*}[h]
    \centering
\includegraphics[width=0.95\columnwidth]{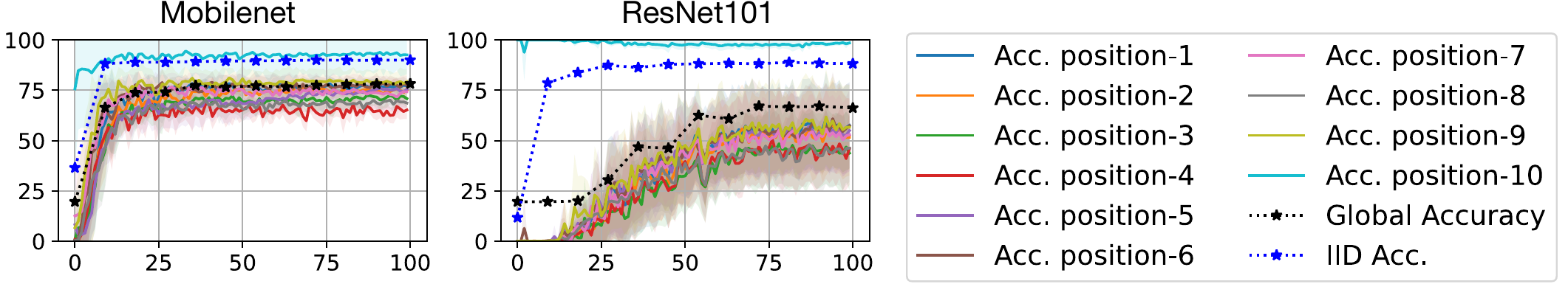}
\caption{\revision{The per-position and global accuracy trained on SVHN with MobileNet and ResNet101. Here we use the shallow-cut and $80\%$-DL. The shaded area around each line represents the confidence interval of $95\%$.}}
\label{fig:svhn_perposition}
\end{figure*}

\textbf{$\bullet$ Figure~\ref{fig:svhn_perposition}:} presents the per-position accuracy for the SVHN dataset trained on the two main models for $80\%$-DL and shallow-cut. This plot is in line with all the aforementioned findings.

\subsection{Impact of the Cut Layer and Comparison with SplitFedv1/FL}
\label{ap:scatter}

\begin{figure*}[!h]
    \centering
\includegraphics[width=1\columnwidth]{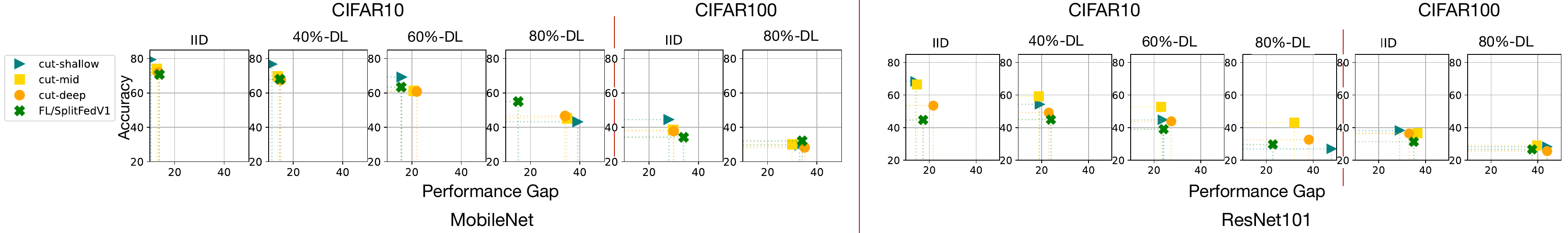}
\caption{Global accuracy~(y-axis) and Performance gap~(x-axis) for SFL~
and FL/SplitFedV1.} 
\label{fig:scatter_plot}
\end{figure*}

Figure~\ref{fig:scatter_plot} presents an extended version of the scatter plot shown in Figure~\ref{fig:per_position_main}, which depicts the global accuracy and the PG for different cut layers (in SFL), and for SplitFedv1~\citep{thapa2022splitfed}, a variant of SL. 
It has been shown that its performance is the same as that of FL~\citep{thapa2022splitfed}, and hence, we denote it by FL. Here we utilize the same cut-layers as in Appendix~\ref{ap:per_pos_all}.  We see
that in terms of accuracy, SFL performs better than FL under IID data. Still, as data heterogeneity
increases, FL performs better. While these results concern cyclic order and DL partition, results on random order and Dirichlet partition are provided in~\citep{hanconvergence}.
Similar are the results for CIFAR-100, even though the performance is not affected as much when the data heterogeneity increases. This is a side-effect of our earlier observation made in Appendix~\ref{ap:per_pos_all} (i.e., lower $L$, present higher intraCF).

\section{Extended Numerical Evaluation}
\label{ap:extend_evaluation}
This part contains supplementary results for the numerical evaluation of Hydra as presented in Section~\ref{sec:exp}. 

\begin{table*}[h]  \caption{PG and global accuracy (reported median of the last five rounds) for cyclic and random order for the SVHN dataset, with $80\%$-DL. This table contains supplementary results of Figure~\ref{fig:medusa_res}. For both models, the small part-2b has been set.
}
\label{tab:hydra_svhn}
\centering 
\begin{tabular}{p{1.5cm}|p{0.9cm}p{0.9cm}p{1.1cm}p{1.1cm}|p{0.9cm}p{1cm}p{1.1cm}p{1.1cm}}

  & \multicolumn{4}{c|}{\textbf{Global Accuracy $\uparrow$}} & \multicolumn{4}{c}{\textbf{Performance gap $\downarrow$}}\\
 &  \small{$\phi=1$} &  \small{$\phi=5$} &  \small{$\phi=10$} & \textbf{random} &  \small{$\phi=1$} &  \small{$\phi=5$} &  \small{$\phi=10$} & \textbf{random}\\ \hline 

&  \multicolumn{8}{c}{ {\textbf{Mobilenet}}} \\ 

SFL  &   {$77.4_{\pm 0.4}$} &  {$77.7_{\pm 0.4}$} &  {$78.2_{\pm 0.5}$} &  {$77.7_{\pm 0.6}$} & {$24_{\pm 0.3}$} &  {$24_{\pm 0.4}$} &  {$21.8_{\pm 0.2}$} &  {$23.8_{\pm 0.7}$}  \\ 

SFL+Hydra &
  {\textbf{$\mathbf{81.3_{\pm 0.1}}$}} &  {\textbf{$\mathbf{81_{\pm 0.1}}$}} &  {\textbf{$\mathbf{81_{\pm 0.1}}$}} &  {\textbf{$\mathbf{82_{\pm 0.3}}$}} &{\textbf{$\mathbf{8.3_{\pm 0.6}}$}} &  {\textbf{$\mathbf{8.3_{\pm 0.5}}$}} &  {\textbf{$\mathbf{10_{\pm 0.4}}$}} &  {\textbf{$\mathbf{9.3_{\pm 0.6}}$}}   
\\
\hline

&  \multicolumn{8}{c}{ {\textbf{ResNet101}}} \\ 

SFL  & 
 {$56_{\pm 5}$} &  {$67_{\pm 3}$} &  {$66.2_{\pm 1.6}$} &  {$59.7_{\pm 2}$} &  {$50_{\pm 3 }$} &  {$46_{\pm 3 }$} &  {$42_{\pm 1}$} &  {$54.9_{\pm 1}$}\\

 SFL+Hydra  &  {\textbf{$\mathbf{75.2_{\pm 1.4}}$}} &  {\textbf{$\mathbf{74.6_{\pm 0.8}}$}} &  {\textbf{$\mathbf{73.4_{\pm 1.5}}$}} &  {\textbf{$\mathbf{66_{\pm 0.9}}$}} & 
 {\textbf{$\mathbf{13_{\pm 0.9}}$}} &  {\textbf{{$\mathbf{13_{\pm 1.5}}$}}} &  {\textbf{$\mathbf{15_{\pm 0.6}}$}} &  {\textbf{$\mathbf{12_{\pm 0.8}}$}} 
\\
\hline
\end{tabular}
\end{table*}

\begin{figure*}[h]
    \centering
\includegraphics[width=0.95\columnwidth]{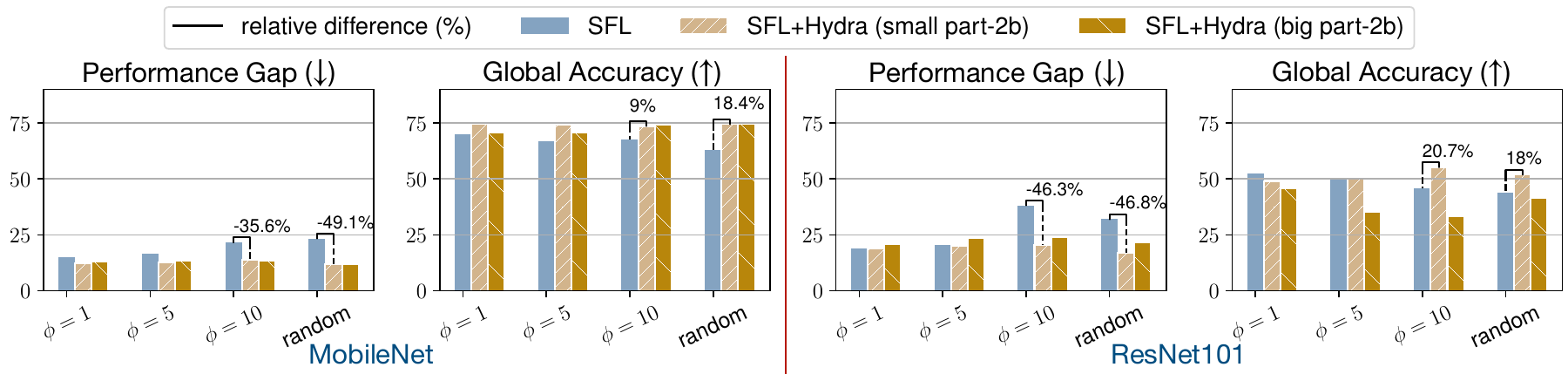}
\caption{The PG and global accuracy (reported median of the last five rounds) using SFL, and SFL+\textit{Hydra} with $60\%$-DL for CIFAR-10 and different processing orders. The over-the-bar text shows the relative difference between SFL and SFL+Hydra.
This Figure contains supplementary results of Figure~\ref{fig:medusa_res} in Section~\ref{sec:exp}.}
\label{fig:medusa_res_app}
\end{figure*}

\begin{figure*}[h]
    \centering
\includegraphics[width=0.95\columnwidth]{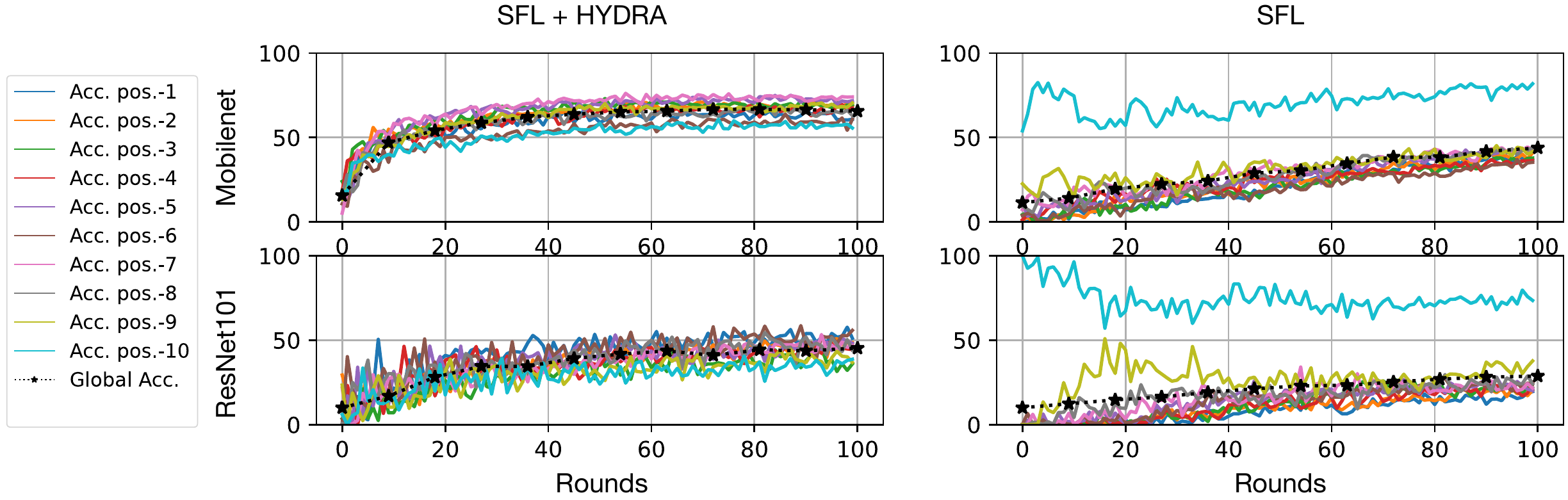}
\caption{\revision{The per-position and global accuracy of SFL VS. SFL+Hydra. The results are for CIFAR-10 with MobileNet and ResNet101. Here we use the shallow-cut and $80\%$-DL, while for Hydra we use the configuration for small part-2b.}}
\label{fig:per_pos_hydra}
\end{figure*}

\begin{table*}[h]  \caption{
Supplementary results of Table~\ref{tab:other_data_pattern}. \\
$\bullet$ \textbf{Top part:} Considering different data partitions and $\phi$ values with SFL+Hydra using MobileNet and CIFAR-10. \\
$\bullet$ \textbf{Bottom part:} Considering different data partitions with SFL+Hydra, and $\phi=10$ for the ResNet101 model and CIFAR-10.
}
\label{tab:other_data_pattern_app}
\centering 
\begin{tabular}{p{2cm}|p{1.1cm}p{1.1cm}p{1.cm}p{1.1cm}|p{1.1cm}p{1.1cm}p{1.cm}p{0.9cm}}
  & \multicolumn{4}{c|}{\textbf{Global Accuracy $\uparrow$}} & \multicolumn{4}{c}{\textbf{Performance Gap $\downarrow$}}\\
  \hline
& \multicolumn{8}{c}{\textbf{MobileNet with CIFAR-10}} \\
\hline
 &  \small{$\phi=1$} &  \small{$\phi=5$}&  \small{$\phi=10$} &  \textbf{random} &  \small{$\phi=1$} &  \small{$\phi=5$} &  \small{$\phi=10$} & \textbf{random}\\ \hline 
\textbf{Sharding}  & $71_{\pm 0.1}$ & $72_{\pm 0.3}$ & $72_{\pm 0.4}$ & $72.5_{\pm 0.4}$ & $12.6_{\pm 0.6}$ & $12.3_{\pm 0.6}$ & $10.8_{\pm 0.4}$ & $12.8_{\pm 1}$\\
\textbf{Dir.-$a_{0.3}$}  & $60_{\pm 0.6}$ & $60_{\pm 0.4}$ & $59_{\pm 0.4}$ & $60_{\pm 0.6}$ & $23.2_{\pm 1}$ & $22.8_{\pm 1}$ & $21_{\pm 1}$ & $20.2_{\pm 2}$\\
 \textbf{Dir.-$a_{0.1}$}  & $49_{\pm 0.2}$ & $49_{\pm 0.5}$ & $49_{\pm 0.2}$ & $49.3_{\pm 0.4}$ & $36.2_{\pm 1}$ & $35_{\pm 0.7}$ & $32_{\pm 0.8}$ & $31.5_{\pm 1}$\\
\hline 
& \multicolumn{8}{c}{\textbf{ResNet101 with CIFAR-10}} \\
\hline
\textbf{Partition Method} & \textbf{Sharding} & \textbf{Dir.$_{\alpha=0.3}$} &  \textbf{Dir.$_{\alpha=0.1}$} & & \textbf{Sharding} & \textbf{Dir.$_{\alpha=0.3}$} &  \textbf{Dir.$_{\alpha=0.1}$} \\ \hline 
SFL+Hydra  & $\mathbf{45_{\pm 1}}$ & $\mathbf{45.7_{\pm 1}}$ & $\mathbf{39_{\pm 0.4}}$ & &  $\mathbf{22_{\pm 2}}$ & $\mathbf{25_{\pm 2}}$ & $\mathbf{30_{\pm 2}}$  \\

SFL &  $42_{\pm 1.2}$ & $33_{\pm 0.6}$ & $27_{\pm 1}$ & &  $35_{\pm 3}$ & $53_{\pm 2}$ & $55_{\pm 3}$ \\
\hline
\end{tabular}
\end{table*}

\subsection{SFL and Hydra}
\label{ap:SFL_VS_HYDRA}
\textbf{Figure~\ref{fig:medusa_res_app}} presents supplementary results of Figure~\ref{fig:medusa_res}. Specifically, it presents the global accuracy and performance gap~(PG) of SFL with and without Hydra~(i.e., SFL and SFL+Hydra) for the two models using CIFAR-10 with $60\%$-DL. For this part, we consider the small part2b for Hydra~(see Section~\ref{sec:exp} for analysis regarding the length of part2b), e.g., the second cut is located at layer $24$, and $33$ for MobileNet, and ResNet101, respectively.  
In detail, Hydra increases the accuracy of SFL with cyclic order up to $9\%$ and drops the corresponding PG up to $35.6\%$. For ResNet101, the accuracy is $20\%$ larger, and the PG is $46.3\%$ smaller compared to regular SFL. Moreover, when considering the random order, Hydra increases the accuracy up to $18\%$ for both ML models. 

Further, \textbf{Figure~\ref{fig:per_pos_hydra} focusing on the per-position} accuracy (median value across all experiments) for $80\%$-DL with CIFAR-10, shows that compared to conventional SFL, SFL+Hydra manages to vanish the gap between the per-position accuracy of position 10 and the rest of the labels.

Moreover, \textbf{Table~\ref{tab:hydra_svhn}} present the performance analysis of models MobileNet and ResNet101 using \textbf{SVHN}. In line with the rest of the results, SFL+Hydra outperforms the conventional SFL training. Specifically, for MobileNet, the drop of PG is up to $65\%$, and the increase in accuracy is up to $5\%$. In fact, we notice that the performance of SFL+Hydra under a high heterogeneity level is very close to the performance of MobileNet under IID~(see Table~\ref{tab:unfairscore_cyclic}). Similarly, for ResNet101, the reduction of the PG is up to $78\%$, while the increase in accuracy is up to $20\%$.

\subsection{Results with Other Types of Ordering}
\label{ap:other_orders_eval}

Figure~\ref{fig:permutation} presents, as well, the performance of SFL+Hydra using cyclic-and-reverse order and small part-2b. This is an alternative structured order as discussed in Appendix~\ref{ap:permutation} and \ref{ap:test_reverse}. The results show that Hydra improves the performance on both the cyclic and cyclic-and-reverse orders. Specifically, Hydra increases accuracy up to $49.7\%$ and decreases PG up to $70\%$ when comparing SFL and SFL+Hydra using cyclic-and-reverse order and the MobileNet using CIFAR-10 with $80\%$-DL. Similarly, for ResNet101 and CIFAR-10 with $80\%$-DL, there is a $72\%$ increase in accuracy and a $75\%$ drop in PG.
In fact, the enhancement of Hydra into SFL for the cyclic-and-reverse order is at the same magnitude as the improvement observed for the cyclic order (see Figure~\ref{fig:medusa_res}).

\subsection{Results with Other Data Partitioning Methods}
\label{ap:other_data_partition}

Table~\ref{tab:other_data_pattern_app} provides additional experiments for Table~\ref{tab:other_data_pattern} in Section~\ref{sec:exp}, which analyzes the performance of Hydra with alternative data partitioning methods; sharding and the Dirichlet method. In detail, the top part of Table~\ref{tab:other_data_pattern_app} shows the results of MobileNet while changing $\phi$, for every alternative data partitioning method. The results show that the performance of SFL+Hydra is not affected by the scale of the clients in all data partitioning methods we study. Hence, this demonstrates that Hydra is robust even in different data partitioning methods. 

Moreover, the second part of the Table provides the results for the ResNet101 model for SFL and SFL+Hydra.  In detail, the results show that for Sharding, Hydra manages a $37\%$ reduction in the PG and a $7\%$ increase in accuracy, when compared with the corresponding values for SFL. While, for Dir.-$\alpha_{0.1}$ Hydra reduces the PG $45\%$, while accuracy increased $44\%$ when compared to regular SFL.

\begin{table*}[t]  \caption{
This table contains supplementary results of Table~\ref{tab:baseline_mobilenet}. The table demonstrates the PG and global accuracy median for the last $5$ rounds across all runs. The SFL+Hydra considers the random processing order, while no service order is required for the others. The
\textbf{(top)} part of the table uses the MobileNet, and \textbf{(bottom)} part uses the ResNet101. Note that the values for CIFAR-100 for some cases are not included in the bottom part. That is because the two corresponding baseline approaches fail to converge, i.e., achieving an accuracy less than $10\%$ after $100$ rounds of training.
}
\label{tab:baseline_mobilenet_app}
\centering 
\begin{tabular}{p{1.7cm}|p{1.4cm}p{1.4cm}p{1.6cm}|p{1.4cm}p{1.4cm}p{1.6cm}}
 & \multicolumn{3}{c|}{\textbf{Global Accuracy $\uparrow$}} & \multicolumn{3}{c}{ \textbf{Performance Gap $\downarrow$}}\\
 & \small \textbf{CIFAR-10} & \small \textbf{CIFAR-10} & \small{\textbf{CIFAR-100}} &\small{\textbf{CIFAR-10}} & \small \textbf{CIFAR-10} & \small \textbf{CIFAR-100} \\ 
 
  & \small{{$80\%$-DL}} & \small {$60\%$-DL} & \small \small {$80\%$-DL} & \small {$80\%$-DL} & \small {$60\%$-DL} &  \small {$80\%$-DL} \\ \hline 
  &  \multicolumn{6}{c}{\textbf{MobileNet}} \\
  \hline
\small SFL+Hydra  & $\mathbf{66.5}$\scriptsize{$\mathbf{\pm 0.1}$} & $\mathbf{74.6{\pm 0.1}}$ & $\mathbf{45_{\pm 0.1}}$ &  $\mathbf{13.4_{\pm 1}}$ & $\mathbf{12_{\pm 1}}$ & \revision{$\mathbf{26.3_{\pm 0.66}}$}\\
\small SplitFedV1/FL & $55_{\pm 0.2}$ & $63.4_{\pm 0.1}$ & $32_{\pm 0.19}$& $15.7_{\pm 1.5}$ & $15{\pm 0.4}$ &  {$34.46_{\pm 0.3}$}\\
\small SplitFedV3 & $43_{\pm 0.3}$ & $51.4_{\pm 0.1}$ & $25.9_{\pm 0.1}$ & $22.6_{\pm 0.8}$ & $22_{\pm 1}$ &  {$47.3_{\pm 0.6}$} \\
\small MultiHead  & $47_{\pm 0.3}$ & $55.7_{\pm 0.46}$ & $25.4_{\pm 0.18}$ & $36.6_{\pm 0.9}$ & $30.5_{\pm 1}$ &  {$59.9_{\pm 0.3}$} \\ \hline 
  &  \multicolumn{6}{c}{\textbf{ResNet101}} \\
  \hline 

  \small SFL + Hydra
  & $\mathbf{49.7_{\pm 1.4}}$ & $\mathbf{51.9_{\pm 0.9}}$ & $\mathbf{33_{\pm 0.9}}$  &  $\mathbf{16.6_{\pm 3.4}}$ &  $\mathbf{17_{\pm2.4}}$ & 
   {$\mathbf{33_{\pm1}}$} \\
\small SplitFedV1/FL 
& $29.7_{\pm 1.8}$ & $39_{\pm 1.1}$ & $23_{\pm 1.6}$ & $22.8_{\pm 2.4}$ & $24.1_{\pm 4}$ & 
 {$43_{\pm 1.6}$} \\
\small SplitFedV3 & $19.1_{\pm 1.6}$ & $35_{\pm 2}$ & -- & $40_{\pm 3}$ & $27.1_{\pm 1}$ & -- \\
\small MultiHead  & $29_{\pm 1}$ & $40_{\pm 0.9}$ & -- & $33_{\pm 2}$ & $28_{\pm 4}$ & --\\
\hline
\end{tabular}
\end{table*}

\subsection{Comparison of Hydra with Baselines and State-of-the-Art Methods}
\label{ap:hydra_vs_baseline}

Table~\ref{tab:baseline_mobilenet_app} presents supplementary results of Table~\ref{tab:baseline_mobilenet} in Section~\ref{sec:exp}. Specifically, it compares the performance of SFL+Hydra using the random order, with additional baseline approaches, such as SplitNN, SplitFedV1/FL, SplitFedV3, and Multihead, as described in Appendices~\ref{ap:protocols}.
The results demonstrate that SFL+Hydra in all cases outperforms all the other benchmarks. Note that for ResNet101, the table does not include results for CIFAR-100 with SplitFedV3 and Multihead. This is because these two approaches fail to converge and effectively complete training. Specifically, after 100 rounds, their accuracy remains below~$10\%$.

\subsection{Weight Regularization}
\label{ap:reg_eval}

\begin{figure*}[h]
    \centering
\includegraphics[width=0.95\columnwidth]{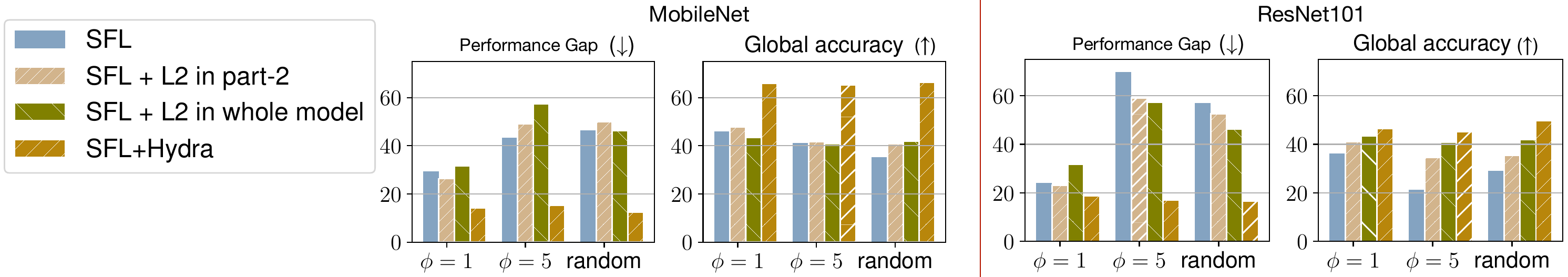}
\caption{Analyzing the \textit{L2-weight Regularization}. The plot shows the average PG and global accuracy for the two models using CIFAR-10, with $80\%$-DL.}
\label{fig:reguralization}
\end{figure*}

As was discussed in Appendix~\ref{ap:reg_CL} the regularization methods used in CL cannot directly be applied in SFL. Nevertheless, we empirically examine L2 regularization, which is a typical approach in ML. Since the server computes the loss function and, by extension, is responsible for adding the regularization term, we consider two cases: \textit{(i)}~computing the regularization term using only the weights of part-2, and \textit{(ii)}~computing the regularization term using the weights of the entire model. Note that the latter case requires additional communication between clients and the server for part-1.  

Figure~\ref{fig:reguralization} presents the result of regular SFL, SFL with L2 regularization, and SFL+Hydra. The plot shows that even with regularization, there is a dependency between the PG and $\phi$, and the type of order~(i.e., the PG gets worse as the system gets more challenging). At the same time, neither of the two types of regularization outperforms the regular SFL in all experiments. For instance, in the case of ResNet101, regularization improves the global accuracy, but the PG increases significantly. Therefore, having just regularization does not guarantee a balance between forgetting and accuracy. On the other hand, SFL+Hydra is the only approach that improves the two scores and remains stable while the $\phi$ increases or when the processing order changes.

\subsection{Ablation Study on the Number of Heads ($G$)}
\label{ap:ablation_study}

In this section, we first provide details about the definition of the superclasses in CIFAR-10 for the results presented in Table~\ref{tab:heads_ablation}.
In particular, similar in spirit to~\citep{bai2021clustering}, we group labels according to their semantic relationships. In detail, for $G=5$, we have defined the groups/superclasses: \textit{land vehicles}: automobile, truck, \textit{things found in the sky}: airplane, bird, \textit{water vehicles}: ship, \textit{domestic animals}: cat, dog, \textit{wild animals}: deer, frog, horse. Furthermore, for the case where $G=2$, we have defined the groups/superclasses \textit{animals}, and \textit{no-animals}, as in~\citep{bai2021clustering}.

Next, we present additional results for CIFAR-100 (trained with MobileNet) in Table~\ref{tab:heads_ablation_app}, where we consider 20 superclasses (of labels) as in~\citep{ramasesh2020anatomy}. We remind the reader that, in Table~\ref{tab:heads_ablation} in the main text, we saw that there is a trade-off between memory reduction and performance when using fewer heads. In particular, even though there is a slight decrease in performance as we use fewer heads, SFL+Hydra still outperforms SFL. Hence, with Hydra, we can reduce the memory demands on the server while still having a better performance than the baseline SFL. On the other hand, in Table~\ref{tab:heads_ablation_app}, for CIFAR-100, we see that the two configurations with $G=20$ and $G=L=100$ have similar performance. This can be justified by an implicit observation derived from Table~\ref{tab:unfairscore_cyclic} in Section~\ref{sec:order}: datasets with a larger number of labels are less sensitive to data heterogeneity and CF in SFL than datasets with a smaller number of labels. In particular, in Table~\ref{tab:unfairscore_cyclic}, we see that the drop in accuracy as the data heterogeneity increases (compared to IID) is larger in CIFAR-10 than in CIFAR-100. It is worth noting that similar observations have been made by related work on FL, e.g., see~\citep{lee2022preservation}. Therefore, the fact that the variations in PG and global accuracy are smaller in Table~\ref{tab:heads_ablation_app} (for CIFAR-100) than the ones in Table~\ref{tab:heads_ablation} (for CIFAR-10) may be related to the observation above.

\begin{table*}[t]  \caption{\revision{PG and global accuracy (of the last five rounds) achieved by MobileNet and CIFAR-100 for cyclic and random order for SFL and Hydra (i.e., SFL+Hydra) with different numbers of heads ($G$). This is a table supplementary to the results shown in Table~\ref{tab:heads_ablation}. }
}
\label{tab:heads_ablation_app}
\centering 
\begin{tabular}{p{1.5cm}p{1.5cm}p{2cm}|p{1.7cm}p{1.3cm}|p{1.7cm}p{1.2cm}}

\textbf{Dataset}& \textbf{Method}& \textbf{Number of} & \multicolumn{2}{c|}{\textbf{Global Accuracy $\uparrow$}} & \multicolumn{2}{c}{\textbf{Performance Gap $\downarrow$}}\\
 & & \textbf{heads} &  \scriptsize{\textbf{Cyclic} $\phi=10$} & \textbf{random} & \scriptsize{\textbf{Cyclic} $\phi=10$} & \textbf{random}\\ \hline  

\multirow[c]{3}{*}{CIFAR-100 } & SFL &    &    $28$\scriptsize{$\pm 0.2$} &  $28.4$\scriptsize{$\pm 0.1$}&  $46$\scriptsize{$\pm 1$} &  $46.7$\scriptsize{$\pm 2$} \\

& SFL+Hydra & $100$   & $\mathbf{33}$\scriptsize{$\mathbf{\pm 0.1}$} &  $\mathbf{33}$\scriptsize{$\mathbf{\pm 0.3}$}&  $\mathbf{45}$\scriptsize{$\mathbf{\pm 1}$}&  $46.5$\scriptsize{$\pm 0.5$} \\

\scriptsize{IID:(33, 42)} & SFL+Hydra & $20$  &   $32$\scriptsize{$\pm 0.1$} &  $33$\scriptsize{$\pm 0.1$} & $46.2$\scriptsize{$\pm 0.6$} &  $\mathbf{46.4}$\scriptsize{$\mathbf{\pm 0.8}$}\\ \hline

\hline
\end{tabular}
\end{table*}

\end{document}